\documentclass[journal,10pt]{IEEEtran}

\usepackage{cite}

\ifCLASSINFOpdf
  \usepackage[pdftex]{graphicx}
  \graphicspath{{figures/}}
  \DeclareGraphicsExtensions{.pdf,.jpeg,.png,.eps}
\else
\fi
\usepackage{xcolor}
\newcommand\hl[1]{\textcolor{black}{#1}}

\usepackage{amsmath}
\usepackage{amsfonts}
\usepackage{amssymb}
\usepackage{bm}
\DeclareMathOperator*{\diag}{diag}

\usepackage{algpseudocode}
\usepackage{algorithm}
\usepackage{booktabs}

\usepackage{array}

\usepackage[flushleft]{threeparttable}
\ifCLASSOPTIONcompsoc
  \usepackage[caption=false,font=normalsize,labelfont=sf,textfont=sf]{subfig}
  \else
  \usepackage[caption=false,font=footnotesize]{subfig}
\fi

\usepackage{url}
\usepackage{hyperref}

\newcommand{\E}{\mathrm{E}}

\newcommand{\Cov}{\mathrm{Cov}}

\begin{document}
\bstctlcite{IEEEexample:BSTcontrol}
\title{Constrained Probabilistic Movement Primitives for Robot Trajectory Adaptation}

\author{Felix~Frank,
        Alexandros~Paraschos,
        Patrick~van~der~Smagt,
        Botond~Cseke%
\thanks{F. Frank, A. Paraschos, P. v.d. Smagt and B. Cseke are with the Volkswagen Machine Learning Research Lab, Munich, 80805 Germany (e-mail: felix.frank@argmax.ai; paraschos@argmax.ai; botond.cseke@argmax.ai; Corresponding author: Felix Frank.)}%
\thanks{A supplementary video can be found at \url{https://youtu.be/7UI6QX-eZ3I}.}}%

\markboth{}%
{Frank \MakeLowercase{\textit{et al.}}: Constrained Probabilistic Movement Primitives for Robot Trajectory Adaptation}

\maketitle

\begin{abstract}
Placing robots outside controlled conditions requires versatile movement representations that allow robots to learn new tasks and adapt them to environmental changes.
The introduction of obstacles or the placement of additional robots in the workspace, the modification of the joint range due to faults or range-of-motion constraints are typical cases where the adaptation capabilities play a key role for safely performing the robot's task.
Probabilistic movement primitives (ProMPs) have been proposed for representing adaptable movement skills, which are modelled as Gaussian distributions over trajectories. These are analytically tractable and can be learned from a small number of demonstrations.
However, both the original ProMP formulation and the subsequent approaches only provide solutions to specific movement adaptation problems, e.g., obstacle avoidance, and a generic, unifying, probabilistic approach to adaptation is missing.
In this paper we develop a generic probabilistic framework for adapting ProMPs.
We unify previous adaptation techniques, for example, various types of obstacle avoidance, via-points, mutual avoidance, in one single framework and combine them to solve complex robotic problems.
Additionally, we derive novel adaptation techniques such as temporally unbound via-points and mutual avoidance.
We formulate adaptation as a constrained optimisation problem where we minimise the Kullback-Leibler divergence between the adapted distribution and the distribution of the original primitive while we constrain the probability mass associated with undesired trajectories to be low.
We demonstrate our approach on several adaptation problems on simulated planar robot arms and 7-DOF Franka-Emika robots in a dual robot arm setting.
\end{abstract}

\IEEEpeerreviewmaketitle

\begingroup
\let\clearpage\relax
\clearpage{}%
\section{Introduction}
\label{sec:intro}
\IEEEPARstart{L}{earning} a robotic task from the ground up requires a considerable effort. The action and sensor spaces are often high-dimensional and the number of repetitions needed to obtain a reasonable representation of these spaces is typically too large for learning them from multiple executions on a real robot.
For improving the learning efficiency we can add prior knowledge, like a model of the environment or a feasible initial solution, to the learning process.
In this paper, we efficiently incorporate prior knowledge through demonstration, relying on humans to show possible solutions to a specific task to a robot, from which it can then generalise.
Learning from demonstration (LfD)~\cite{calinonLearningDemonstrationProgramming2018} frameworks have been proposed and, typically, they break down complex tasks into simple movements, called movement primitives, capable of being learned from a few demonstrations and subsequently combined in order to solve more complex tasks.

There are approaches to movement primitives that use a \emph{deterministic} representation of trajectories, such as dynamic movement primitives (DMP) as formulated in \cite{ijspeertLearningAttractorLandscapes2003,ijspeertDynamicalMovementPrimitives2012} or central pattern generators (CPG)~\cite{ijspeertCentralPatternGenerators2008}, which learn a deterministic average of the given demonstrations.
In addition, approaches with a \emph{probabilistic} trajectory representation, such as probabilistic movement primitives (ProMP)~\cite{paraschosProbabilisticMovementPrimitives2013}, probabilistic formulations of dynamic movement primitives \cite{meierProbabilisticRepresentationDynamic2016,calinonStatisticalDynamicalSystems2012}, Gaussian mixture models with Gaussian mixture regression (GMM-GMR)~\cite{calinonLearningRepresentingGeneralizing2007} or kernelized movement primitives (KMP)~\cite{huangKernelizedMovementPrimitives2019}, learn the inherent variability in the demonstrations.

Probabilistic approaches are able to represent both the mean and variance of multiple demonstrated trajectories.
Some methods directly model trajectories in the demonstrated space with Gaussian mixture models~\cite{calinonLearningRepresentingGeneralizing2007}, whereas other methods learn the parameters of a mixture of linear Gaussian state-space models~\cite{chiappaUsingBayesianDynamical2009} to best represent the given trajectories.
Probabilistic movement primitives (ProMPs) represent the distribution over trajectories by a linear combination of a stochastic weight vector and a set of basis functions~\cite{paraschosProbabilisticMovementPrimitives2013}.

In practice there is a wide variety of potential tasks and it is impractical to demonstrate every possible variation to the robot beforehand.
\hl{Therefore, a key element in every movement primitive framework is the ability to adapt primitives to unseen scenarios.}
\hl{For adaptation, probabilistic approaches can utilise the additional information encoded in the covariance by adapting the trajectory only at the points of interest while the remainder of the trajectory remains largely unchanged.
The remaining variability can then be exploited further downstream, for instance to satisfy further constraints or to optimise some criterion.
Since approaches with deterministic primitive representations do not encode this variability, adapting them, for example, by adding a via-point, can only be done by ad-hoc methods.}

\begin{table*}[t]
  \caption{Feature comparison of adaptation techniques for movement primitive frameworks}
  {\centering%
  \begin{tabular}{l>{\centering}p{9mm}>{\centering}p{12mm}>{\centering}p{11mm}>{\centering}p{8mm}>{\centering}p{6mm}>{\centering}p{10mm}>{\centering}p{10mm}>{\centering}p{10mm}>{\centering}p{12mm}>{\centering\arraybackslash}p{10mm}}
  \toprule
  Feature & Proba-bilistic & Point obstacles & Volumetric obstacles & Virtual walls & Joint \newline limits & Smoothness & Via-points\footnotemark & Space transformation & Temporally unbound waypoints & Mutual avoidance \\
  \midrule
  Ours & \checkmark & \checkmark & \checkmark & \checkmark & \checkmark & \checkmark & \checkmark & \checkmark & \checkmark & \checkmark\\
  \midrule
  Previous work on ProMPs\\
  \cmidrule{1-1}
  Paraschos et al.~\cite{paraschosProbabilisticMovementPrimitives2013,paraschosProbabilisticPrioritizationMovement2017} & \checkmark & \textbf{-} & \textbf{-} & \textbf{-} & \textbf{-} & \checkmark & \checkmark & \checkmark & \textbf{-} & \textbf{-}\\
  Koert et al. \cite{koertDemonstrationBasedTrajectory2016} & \checkmark & \checkmark & \checkmark & \textbf{-} & \textbf{-} & \textbf{-} & \checkmark & \textbf{-}\footnotemark & \textbf{-} & \textbf{-}\\
  Colome et al. \cite{colomeDemonstrationfreeContextualizedProbabilistic2017} & \textbf{-} & \checkmark & \textbf{-} & \textbf{-} & \textbf{-} & \textbf{-} & \checkmark & \textbf{-} & \textbf{-} & \textbf{-}\\
  Koert et al. \cite{koertLearningIntentionAware2019} & \textbf{-} & \checkmark & \textbf{-} & \textbf{-} & \textbf{-} & \textbf{-} & \checkmark & \textbf{-} & \textbf{-} & \textbf{-}\\
  Gomez et al. \cite{gomez-gonzalezUsingProbabilisticMovement2016,gomez-gonzalezAdaptationRobustLearning2020}& \checkmark & \textbf{-} & \textbf{-} & \textbf{-} & \textbf{-} & \textbf{-} & \checkmark & \checkmark & \textbf{-} & \textbf{-}\\
  Osa et al.~\cite{osaGuidingTrajectoryOptimization2017} & \textbf{-} & \checkmark & \textbf{-} & \textbf{-} & \textbf{-} & \checkmark & \checkmark & \checkmark & \textbf{-} & \textbf{-}\\
  Shyam et al.~\cite{shyamImprovingLocalTrajectory2019} & \textbf{-} & \checkmark & \textbf{-} & \textbf{-} & \textbf{-} & \checkmark & \checkmark & \checkmark & \textbf{-} & \textbf{-}\\
  \midrule
  Other MP frameworks\\
  \cmidrule{1-1}
  DMP~\cite{ijspeertLearningAttractorLandscapes2003} & \cite{calinonStatisticalDynamicalSystems2012,meierProbabilisticRepresentationDynamic2016} & \cite{parkMovementReproductionObstacle2008,hoffmannBiologicallyinspiredDynamicalSystems2009,chiLearningGeneralizationObstacle2019,krugModelPredictiveMotion2015} & \cite{ginesiDynamicMovementPrimitives2019} & \cite{saverianoLearningBarrierFunctions2019} & \cite{duanConstrainedDMPsFeasible2018} & \textbf{-} & \checkmark & \textbf{-} & \textbf{-} & \textbf{-}\\
  GMM-GMR \cite{calinonLearningRepresentingGeneralizing2007} & \checkmark & \cite{huangGeneralizedTaskParameterizedSkill2018} & \cite{huangGeneralizedTaskParameterizedSkill2018} & \textbf{-} & \textbf{-} & \textbf{-} & \checkmark & \checkmark & \textbf{-} & \textbf{-}\\
  KMP~\cite{huangKernelizedMovementPrimitives2019} & \checkmark & \cite{huangEKMPGeneralizedImitation2021} & \cite{huangEKMPGeneralizedImitation2021} & \cite{huangLinearlyConstrainedNonparametric2020} & \cite{huangLinearlyConstrainedNonparametric2020} & \cite{huangOrientationLearningAdaptation2020} & \checkmark & \textbf{-} & \textbf{-} & \textbf{-}\\
  \bottomrule
  \vspace*{1mm}
  \end{tabular}
  \par}
  {\footnotesize
  We consider a method as \textbf{probabilistic} if it retains a probabilistic representation after adaptation.
  We compare different approaches on whether they can adapt primitives to avoid \textbf{point obstacles}, \textbf{virtual walls}, \textbf{joint limits} or \textbf{volumetric obstacles}, which we consider as spherical objects with a specified radius.
  We also consider approaches that can influence the \textbf{smoothness} of a primitive or force a primitive to pass through specified \textbf{via-points}.
  \textbf{Space transformations} allow the adaptation of a joint space primitive with requirements from a different space, commonly that is the Cartesian space.
  In this paper we develop two novel adaptation techniques: \textbf{Temporally unbound waypoints} enforce via-points that have to be passed through at an unspecified time during the primitive, whereas \textbf{mutual avoidance} allows us to combine primitives of different robots within the same Cartesian space while avoiding collisions between robots. 
  }
  \label{tab:lit_feature_matrix}%
\end{table*}

There is extensive literature on adapting the different primitive frameworks to specific tasks.
The DMP formulation has been extended to include obstacle avoidance~\cite{parkMovementReproductionObstacle2008,hoffmannBiologicallyinspiredDynamicalSystems2009,chiLearningGeneralizationObstacle2019,krugModelPredictiveMotion2015,ginesiDynamicMovementPrimitives2019}, as well as joint limits~\cite{duanConstrainedDMPsFeasible2018} and limits on the robot velocity~\cite{dahlinAdaptiveTrajectoryGeneration2020}.
In the probabilistic frameworks one direction of research combines multiple primitives together in one primitive~\cite{calinonHandlingMultipleConstraints2009,calinonStatisticalLearningImitation2009,calinonTutorialTaskparameterizedMovement2016,paraschosProbabilisticPrioritizationMovement2017,silverioLearningTaskPriorities2019}, allowing adaptation by \emph{including} additional behaviour into a primitive.
Conversely, robotic problems like avoidance or respecting joint limits are typically easier to formulate by \emph{excluding} specific behaviour from a primitive.
In this direction, extensions to the ProMP framework for obstacle avoidance have been investigated in~\cite{koertDemonstrationBasedTrajectory2016,colomeDemonstrationfreeContextualizedProbabilistic2017,koertLearningIntentionAware2019}, but their approaches present specific solutions and a probabilistic approach tackling primitive adaptation in a generic way is currently missing.
Furthermore, a combination of these methods is often not possible because many of them optimise for deterministic trajectories and loose the probabilistic description of the primitive.

Our paper has two core contributions.
First, we provide a generic and unifying formulation for \emph{probabilistic} adaptations of ProMPs where all the proposed adaptation techniques can be combined in a principled way to solve complex robotic problems.
Retaining the probabilistic description after adaptation is a major difference to many of the current state-of-the-art approaches~\cite{koertLearningIntentionAware2019,colomeDemonstrationfreeContextualizedProbabilistic2017,osaGuidingTrajectoryOptimization2017,shyamImprovingLocalTrajectory2019}.
This allows building libraries of probabilistic primitives and enables further downstream \emph{probabilistic} adaptation or co-activation of primitives.
Second, we develop \emph{probabilistic} adaptation techniques novel to the ProMP framework, i.e., virtual walls and joint limits, as well as temporally unbound waypoints and dual-arm avoidance. 
The latter two, to our knowledge, are novel not only to the ProMP framework, but they have not been presented in any primitive framework before.

We formulate primitive adaptation as a constrained optimisation problem, in which the goal is to stay close to a given initial primitive, while fulfilling a set of constraints.
Throughout the adaptation we retain the probabilistic characteristics and we synthesise a new probabilistic primitive (ProMP) that can be reproduced on a robot using the known ProMP framework.
In Table~\ref{tab:lit_feature_matrix} we compare the adaptation techniques our method offers to previous approaches on ProMP adaptation and the adaptation techniques available in other movement primitive frameworks.

The paper is structured as follows. In Section~\ref{sec:method} we give a short introduction to probabilistic movement primitives, followed by the problem formulation.
Afterwards, we develop multiple adaptation techniques using probabilistic constraints.
Section~\ref{sec:experiments} presents several applications of our method and we compare our method to state-of-the-art approaches in Section~\ref{sec:related}.
\footnotetext[1]{We grant all ProMP based approaches this feature, because the conditioning mechanism is inherent to ProMPs, however, it should be noted that our approach offers specifying via-points with an allowed radius, whereas conditioning requires tuning the variance to realise a desired tolerance.}
\footnotetext[2]{This feature could be developed with a straightforward extension.}
\clearpage{}%

\clearpage{}%
\section{Methodology}
\label{sec:method}
In this section we derive our adaptation framework for ProMPs and we propose several constraints to adapt primitives to novel situations.
We begin by providing background on ProMPs. 

\subsection{Background on Probabilistic Movement Primitives}
\label{SecBackgroundProMP}
Probabilistic Movement Primitives (ProMPs)\cite{paraschosProbabilisticMovementPrimitives2013} are a method to model trajectory distributions that are extensively used in robotics~\cite{paraschosUsingProbabilisticMovement2018,gomez-gonzalezUsingProbabilisticMovement2016,gomez-gonzalezAdaptationRobustLearning2020,koertDemonstrationBasedTrajectory2016,koertLearningIntentionAware2019,colomeDemonstrationfreeContextualizedProbabilistic2017,lioutikovLearningMovementPrimitive2017,maedaProbabilisticMovementPrimitives2017}. A trajectory $\bm{y}=\{\bm{y}_{t_l}\}^{L}_{l=0}$ typically consists of recordings of a robot's joint space or certain Cartesian coordinates $\bm{y}_t   \in \mathbb{R}^{D}$ at times $t_l \in [0,T]$. In this paper we develop models and methods for ProMPs with joint-space coordinates. Applying them to ProMPs modelling Cartesian coordinates is straightforward and can be viewed as a simpler special case---we point out the differences at the appropriate places.

The coordinates $\bm{y}_t$ are modelled as a linear combination $\bm{z}_t(\bm{w})=\bm{w} \bm{\phi}_t$ 
under the presence of zero-mean Gaussian noise, that is,
\begin{equation}
\bm{y}_{t_l} = \bm{z}_{t_l}(\bm{w}) + \epsilon_{t_l},
\quad  \epsilon_{t_l} \sim \mathcal{N}\left(0, \bm{\Sigma}_y\right).
\end{equation}
The vector $\bm{\phi}_t = [\phi^1(t), \phi^2(t), \ldots, \phi^M(t)] \in \mathbb{R}^M$ consists of the values at time $t$ of $M$ basis functions,  whereas $\bm{w} \in \mathbb{R}^{D \times M}$ is a stochastic weight matrix with the same number of rows $D$ as $\bm{y}_t$. As a result, the observation model is given by
\begin{equation}\label{dist_timepoint}
p(\bm{y}_{t} \vert \bm{w}) = \mathcal{N}(\bm{y}_{t} ; \bm{w}\bm{\phi}_{t}, \bm{\Sigma}_y).
\end{equation}
The distribution over weights $p(\bm{w})$ is chosen to be Gaussian, i.e.,  $p(\bm{w}) =  \mathcal{N}(\mathrm{vect}(\bm{w}) ; \mathrm{vect}(\bm{\mu}_w), \bm{\Sigma}_w)$, where $\mathrm{vect}(\bm{w})$ and $\mathrm{vect}(\bm{\mu}_w)$ are blocked column vectors w.r.t. the rows of
$\bm{w}$ and $\bm{\mu}_{w}$, respectively. For reasons of notational brevity, in the following we use 
$ p(\bm{w}) =  \mathcal{N}(\bm{w} ; \bm{\mu}_w, \bm{\Sigma}_w)$.
The marginal distribution of a trajectory $\bm{y}_{0:T} $ can thus be written as
\begin{equation}\label{dist_traj}
p(\bm{y}_{0:T} \vert \bm{\theta}) = \int \! \mathcal{N}(\bm{w} ; \bm{\mu}_w, \bm{\Sigma}_w) \prod_{l=1}^L \mathcal{N}(\bm{y}_{t_l} ; \bm{w} \bm{\phi}_{t_l}, \bm{\Sigma}_y)  d\bm{w},
\end{equation}
where $\bm{\theta} = \{ \bm{\mu}_w, \bm{\Sigma}_w, \bm{\Sigma}_y \}$ denotes all learnable model parameters.

In robotics, one typically records joint-space position and velocity at discrete points in time.
\hl{Position and velocity can be modelled either as independent variables or using the same weight vector and dependent basis functions, that is, the pair $[\phi_t, \dot{\phi}_t]$ maps to position and velocity respectively~\cite{paraschosProbabilisticMovementPrimitives2013}.}
The basis functions are commonly chosen as Gaussian radial basis functions $\phi^i_t = \exp(-(\tau(t) - c_i)^2 / (2h))$, where the phase variable $\tau(t)$ is a monotone mapping from recorded time to the interval $[0, 1]$.
Scaling time to a common phase space enables using trajectories of different length in one primitive and the phase variable allows the adaptation of the execution speed.
The radial basis functions are chosen because they are localised in time and are infinitely differentiable.

Given a set of trajectories $\{\bm{y}^i\}^n_{i=1}$, which are considered to be  i.i.d., one can learn the distribution over weights by maximum likelihood estimation, i.e.,
\begin{align}\label{EqnML}
 \bm{\theta}^{\ast} = \mathop{\mathrm{argmax}}_\theta 
\sum_i \log p(\bm{y}_{0:T}^{i} \vert \bm{\theta}). 
\end{align}
The maximisation can be carried out with the \emph{Expectation Maximisation} (EM) algorithm. EM iterates through the E- and M-steps until convergence is achieved. In the E-step the individual posteriors $p(\bm{w} \vert \bm{y}^i, \theta_{s}) \propto p(\bm{y}^i \vert \bm{w}, \theta_{s}) \, p(\bm{w} \vert \theta_s)$ are computed, while in the M-step a Gaussian $p(\bm{w} | \theta_{s+1})$ that matches the moments of $\frac{1}{n}\sum_i p(\bm{w} \vert \bm{y}^i, \theta_{s})$ is fitted to update the parameters $\bm{\theta}$.
In many practical applications it is useful to add an additional penalty term $\log p(\bm{\theta})$ to the objective in \eqref{EqnML} and thus resort to maximum a posteriori estimation. A common choice for $p(\bm{\theta})$ is $p(\bm{\theta}) = p(\bm{\mu}_w)\,p(\bm{\Sigma}_w)\,p(\bm{\Sigma}_y)$ with Gaussian  $p(\bm{\mu}_w)$ and inverse-Wishart  $p(\bm{\Sigma}_w)$ and $p(\bm{\Sigma}_y)$, as in \cite{paraschosUsingProbabilisticMovement2018} and \cite{gomez-gonzalezAdaptationRobustLearning2020}.
This regularisation can be particularly important if a subsequent adaptation requires a more flexible function class than the (original) ProMP would %
and thus one runs the danger of overfitting the data when learning the (original) ProMP.  

The distribution $p(\bm{w} \vert \theta^{\ast})$ encodes the probabilistic model learnt from the set of trajectories  $\{\bm{y}^i\}^n_{i=1}$, thereby capturing their inherent variability.
As a result, ProMPs can be used to generate trajectories similar to $\{\bm{y}^i\}^n_{i=1}$ through sampling. Moreover, the stochastic ProMP controller allows us to reproduce these trajectories on a physical system~\cite{paraschosProbabilisticMovementPrimitives2013}.
In the context of learning from demonstrations, the trajectories $\{\bm{y}^i\}^n_{i=1}$ are typically provided by a human expert.

An important aspect of modelling with ProMPs is adapting the learned primitives to new situations. 
Previous approaches use, for example, Gaussian conditioning to incorporate via- or end-points in joint space \cite{paraschosProbabilisticMovementPrimitives2013} or task space \cite{gomez-gonzalezAdaptationRobustLearning2020}. Other features include merging different primitives\cite{paraschosProbabilisticMovementPrimitives2013} or incorporating obstacle avoidance into a trajectory\cite{koertDemonstrationBasedTrajectory2016,colomeDemonstrationfreeContextualizedProbabilistic2017,koertLearningIntentionAware2019}.
In the following we develop our unified framework for ProMP adaptation, which is based on constraining the probability mass associated with undesirable trajectories to be~low.

\subsection{Adaptation by Trajectory Constraints}
Due to the Gaussian choice of the ProMP distribution $p(\bm{w})$ the trajectory $\{\bm{z}_t\}, \: t \in [0,T]$ is a Gaussian process with mean value function $\E[ \bm{z}_t ] = \bm{\mu}_w \bm{\phi}_t$ and covariance function $\Cov [z^i_t, z^j_{t^\prime} ] = \bm{\phi}_{t}^T \bm{\Sigma}_w^{(i,j)} \bm{\phi}_{t^\prime}$, where $\bm{\Sigma}_w^{(i,j)}$ is the covariance matrix block between the rows $\bm{w}_i$ and $\bm{w}_j$ of $\bm{w}$. Therefore, any finite dimensional joint distribution of $\{ \bm{z}_t \}$ is also Gaussian and can be easily computed.
We define $\{\bm{x}_t\}$, with $\bm{x}_t(\bm{w}) \equiv \bm{T}(\bm{z}_t(\bm{w}))$, as the process corresponding to certain Cartesian coordinates of interest, such as end-effector or elbow pose or velocity.
The function $\bm{T}$ maps the robot's joint state $\bm{z}_t$ to the Cartesian coordinates $\bm{x}_t$ and it is commonly referred to as the forward kinematics function~\cite{sicilianoSpringerHandbookRobotics2016}. When modelling in the Cartesian space, we simply set $\bm{T}$ to the identity function, that is,~ $\bm{x}_t(\bm{w}) \equiv \bm{z}_t(\bm{w})$.

Generally, in any given robotics task, we have a set of points of interest $\bm{x}^{k}_t(\bm{w})$, for example the robot's end effector and its elbow joint, and corresponding forward kinematics functions $\bm{T}_k$.
Without loss of generality and for reasons of notational brevity, we limit our presentation to a single point of interest~$\bm{x}_t(\bm{w})$.  

Since $\{\bm{z}_t\}$ is a well defined stochastic process, we can express trajectory constraints in terms of probabilities.
That is, we can associate a probability mass to a trajectory constraint $c(\{\bm{z}_t(\bm{w})\}) \leq d$ being valid.
For example, an end-point constraint  $\bm{z}_T(\bm{w}) \leq d$ being valid with confidence $\alpha$ can be formulated as $P_{\bm{w}}( \bm{z}_T(\bm{w}) \le d) \ge \alpha$. Here $c(\{\bm{z}_t(\bm{w})\}) = \bm{z}_T(\bm{w}) $ and $P_{\bm{w}}$ denotes the probability measure corresponding to the density~$p(\bm{w})$.

In this paper we consider two types of constraints. First \emph{point constraints} that take as input $\bm{x}_t$ or $\bm{z}_t$, such as via- or end-point constraints, and second \emph{path constraints} that are applied to the path $\{\bm{x}_t\}_{t\in \mathcal{T}}$ or $\{\bm{z}_t\}_{t\in \mathcal{T}}$ with contiguous temporal support $\mathcal{T}\subseteq [0,T]$.
For example, bounding the smoothness of a trajectory can be considered a \emph{path constraint}.
Generally, we formulate $K$ \emph{point constraints} as inequality constraints $P_{\bm{w}}\big(c_{k,t}(\bm{x}_t(\bm{w})) \le d_{k,t}\big) \ge \alpha_{k,t}$ that are required to be valid at time-point $t$, whereas \emph{path constraints} are formulated as $P_{\bm{w}}\big(c_{k}(\{ \bm{x}_t(\bm{w}) \}_{t\in \mathcal{T}} ) \le d_{k}\big) \ge \alpha_{k}, \:  k \in \{1, \ldots, K\}$. While not all these quantities are analytically tractable, for the constraints we consider in this paper we provide accurate approximations that lead to state-of-the-art results.
In the following derivation of the problem statement we use $c_{k,t}$ as a placeholder for the specific constraints which are developed in the Sections~\ref{sec:method_jconst} and~\ref{sec:method_task}.

\subsection{Problem Formulation}
\label{SecMethProblem}
ProMPs represent the information learnt from the trajectories $\{\bm{y}^i\}^n_{i=1}$ with a Gaussian distribution which we now denote by $p_0(\bm{w}) =\mathcal{N}(\bm{w} ; \bm{\mu}_{w}^0, \bm{\Sigma}_{w}^0)$.
We view adaptation as imposing constraints on the paths generated by this distribution.
In this view, adaptation means computing a new $p(\bm{w})$ which is as close as possible to $p_0(\bm{w})$, while paths generated from $p(\bm{w})$ satisfy a set of constraints. We formulate the adaptation of ProMPs as the constrained optimisation problem,
\begin{subequations}
\begin{align}
  \min_{p} \: & \: D_{\textrm{\sc KL}}[\,p(\bm{w}) \vert\!\vert\, p_0(\bm{w})] \label{pf_kl2promp} \\
  \text{s.t} \: & \:\E_{p(\bm{w})}[H(d_{k,t} - c_{k,t}(\bm{w}))] \ge \alpha_{k,t} \quad \forall \: k,t \in \mathcal{T}_k, \label{pf_const}
\end{align}
\end{subequations}
where for convenience we write the probability mass constraints as expectation constraints
\begin{equation}
 \E_{p(\bm{w})}[H(d_{k,t} - c_{k,t}(\bm{w}))] =  P_{\bm{w}}\big(c_{k,t}(\bm{w}) \leq d_{k,t} \big).
\end{equation}
The function $H$ stands for the heavyside step function and the set $\mathcal{T}_k$ denotes the temporal support of the $k^{th}$ constraint. 
For example, for an end-point constraint we have $\mathcal{T}_k=\{T\}$, while when limiting the end-effector trajectories to a predefined area for a certain time we have $\mathcal{T}_k \subseteq [0,T]$
We visualise different temporal supports of constraints in Figure~\ref{fig:joint_smooth}.
The objective in Equation (\ref{pf_kl2promp}) is the (reverse) Kullback-Leibler divergence.
We choose this divergence because of its analytical tractability w.r.t. the Gaussian family of distributions and because of its mode-seeking property \cite{minkaDivergenceMeasuresMessage2005} when compared to $ D_{\textrm{\sc KL}}[\,p_0(\bm{w}) \vert\!\vert\, p(\bm{w})]$.
The latter is particularly important because the distributions carved out by the constraints are often multimodal.
In this formulation, without lack of generality, we omit the dependence of $c_{k,t}$ on either $\bm{x}_t(\bm{w})$ or $\bm{z}_t(\bm{w})$ because all the stochasticity in the model is captured by the variable $\bm{w}$.
Note that there should be an additional normalisation constraint for $p$, which we omit for notational brevity.
Furthermore, \emph{path constraints} can be added by imposing~$\E_{p(\bm{w})}[H(d_{k} - c_{k}(\bm{w}))] \ge \alpha_k$. 

Alternatively, instead of imposing strict limits, we can relax constraints and recast them as penalties which we add to the objective  \eqref{pf_kl2promp}.  For example, adding a smoothness regulariser $R(\{\bm{z}_t(\bm{w})\})$ leads to a trade-off of staying close to $p_0(\bm{w})$ and prioritising smooth trajectories. To achieve this, we can add $\kappa \,\E_{p(\bm{w})}[R(\bm{w})]$ to the objective~\eqref{pf_kl2promp}.

The constrained optimisation problem in (\ref{pf_kl2promp}) and (\ref{pf_const}) is a convex problem with linear constraints w.r.t. $p(\bm{w})$. The corresponding Lagrangian reads as
{%
\begin{align}
\label{pfLagrangeP}
  L(p,\{\lambda_{k,t}\})=& D_{\textrm{\sc KL}}[\,p(\bm{w}) \vert\!\vert\, p_0(\bm{w})]\\
  & + \sum_{k,t} \lambda_{k,t}\: \big[\alpha_{k,t} - \E_{p(\bm{w})}[H(d_{k,t} - c_{k,t}(\bm{w}))]\big] \nonumber
\end{align}
}
and the resulting optimisation problem is formulated as
\begin{align}
\label{pf_dual}
  \max_{\lambda_{k,t} \geq 0} \min_{p} L(p, \{\lambda_{k,t}\}).
\end{align}
The optimisation problem is convex-concave in $p$ and $\{\lambda_{k,t}\}$ and the optimal $(p^\ast, \{\lambda^{\ast}_{k,t}\})$ satisfies
\begin{align}
	\label{EqnOptimalSolution}
	p^{\ast}(\bm{w}) \propto p_0(\bm{w}) \exp\Big\{ \sum_{k,t} \lambda_{k,t}^{\ast} H(d_{k,t} - c_{k,t}(\bm{w})) \Big\}.
\end{align} 
This distribution is non Gaussian, non-smooth, and often multimodal. However, we are interested in finding a Gaussian $p^{\ast}(\bm{w})$ to be able to keep the ProMP formulation and, subsequently, its properties. Additionally, keeping the ProMP formulation allows us to use a known stochastic controller that can reproduce the trajectory distribution on a physical system \cite{paraschosProbabilisticMovementPrimitives2013}. For this reason, in the following we restrict the search space of $p(\bm{w})$ to Gaussians.

Let $p(\bm{w}) = \mathcal{N}(\bm{w} ; \bm{\mu}_{w}, \bm{\Sigma}_{w})$ and let $F_{c_{k,t}}(\cdot ; \bm{\mu}_{w}, \bm{\Sigma}_{w})$ denote the cumulative distribution function of the random variable defined by $c_{k,t}(\bm{w})$.  
\hl{
    With this parameterisation the constraints are typically no longer convex, however, for the problems we formulate in this paper the resulting optimal solutions can be found by gradient ascent-descent methods. 
}
The Lagrangian resulting from this parameterisation can be written~as
\begin{align}
  L(\bm{\mu}_w, \bm{\Sigma}_w, & \{\lambda_{k,t}\}) =  D_{\textrm{\sc KL}}[\,\mathcal{N}(\bm{w} ; \bm{\mu}_{w}, \bm{\Sigma}_{w}) \vert\!\vert\, \mathcal{N}(\bm{w} ; \bm{\mu}_{w}^0, \bm{\Sigma}_{w}^0)]
  \nonumber
   \\
  & + \sum_{k,t} \lambda_{k,t}\: \big[\alpha_{k,t} - F_{c_{k,t}}(d_{k,t} ; \bm{\mu}_{w}, \bm{\Sigma}_{w}) \big]. 
  \label{pf_loss}
\end{align}
In the following we use this formulation to adapt ProMPs to new scenarios.
We formulate every adaptation by defining the corresponding $c_{k,t}(\bm{w})$ and approximating its cumulative distribution function $F_{c_{k,t}}$.
Moreover, we use differentiable approximations so that the gradients of the Lagrangian can be numerically computed using automatic differentiation frameworks.

\begin{figure}[t]
  \centering
  \includegraphics[width=0.95\columnwidth]{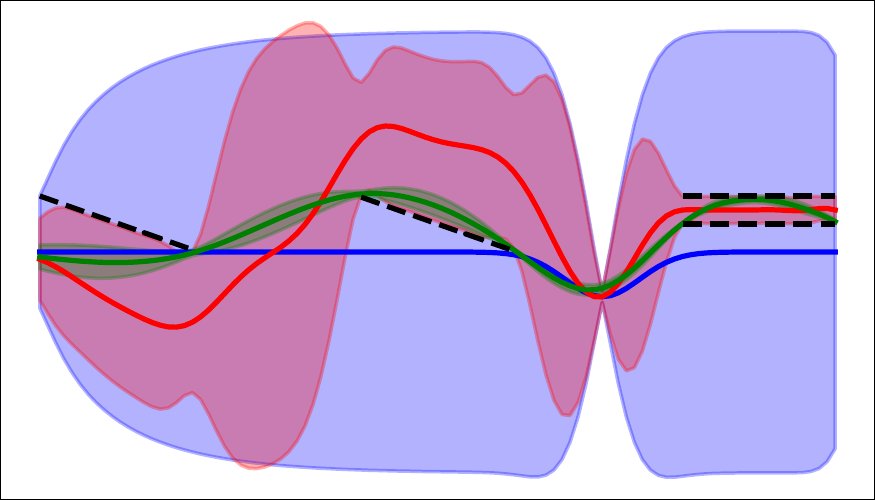}
  \caption{A toy example illustrating limit constraints (dashed, black lines) and the effects of the smoothness regulariser introduced in Section~\ref{sec:method_jconst}.
  The wide Gaussian prior (blue) is adapted to obey limit constraints (red).
  A smoother distribution of trajectories with lower variance is obtained when a smoothness regulariser ($\kappa=0.01$) is added (green).
  The thick line corresponds to the mean of the respective primitive, whereas the shaded area indicates three standard deviations.}
  \label{fig:joint_smooth}
\end{figure}

\subsection{Joint space constraints}
\label{sec:method_jconst}
These constraints depend directly on $\{ \bm{z}_{t}(\bm{w})\}$ and hence their computation does not involve an often non-linear forward kinematics function.

\subsubsection{Limits on joint space coordinates}
\label{sec:method_jconst_limits}
In general, every robotic movement is limited to a specific motion range depending on the robot's links and their configuration.
These limits can also change during operation, for example, we might want to limit the range of motion of a link after a failure, or when an added sensor blocks part of the motion range.
Changing these limits requires an adaptation of the learned movement primitives.  
For dynamic movement primitives, the problem of joint-limit avoidance has been addressed by learning primitives in a transformed space~\cite{duanConstrainedDMPsFeasible2018}.
However, this method requires demonstrations to already conform to the limits, otherwise a mapping of demonstrations to the transformed space is not straightforward.  
In our probabilistic framework, we can add constraints to an existing trajectory distribution.
A one-sided constraint can be formulated~as  
\begin{align}
P_{\bm{w}} ( z^k_t(\bm{w}) \le b_{k,t}) \ge \alpha_{k,t},
\end{align}
while for a two-sided constraint we can~use
\begin{align}
P_{\bm{w}} (a_{k,t} < z^k_t(\bm{w}) \le b_{k,t}) \ge \alpha_{k,t}.
\end{align}
In this case, instead of the step function $H(\cdot)$, we can use the indicator function $I_{D_{k,t}}(\cdot)$ of the interval $D_{k,t} = [a_{k,t}, b_{k,t}]$ to formulate the probability mass constraint as an expectation constraint.
The latter can then be added to \eqref{pf_const} by using 
\begin{align}
\E[{I}_{D_{k,t}} (z^k_t)] = \Phi\left[\frac{b_{k,t} - \E[z^k_t]}{V[z^k_t]^{1/2}} \right] - \Phi\left[ \frac{a_{k,t} - \E[z^k_t]}{V[z^k_t]^{1/2}} \right],
\end{align}
with $\Phi$ denoting the CDF of a standard normal random variable.
Note that these constraints only satisfy the joint limits in the probabilistic sense, that is, the probability of being out of bounds is very low.
Nonetheless, this method can be used to adapt to changing operational conditions, whereas for simple joint limit avoidance one should rely on learning the movement primitives in a transformed space, analogous to \cite{duanConstrainedDMPsFeasible2018}, resulting in stronger guarantees.

In Figure~\ref{fig:joint_smooth}, we illustrate a toy adaptation problem for a univariate ProMP with time-varying one-sided and fixed two sided limit constraints.
The original ProMP is a wide non-informative Gaussian with a via-point computed by conditioning.
The adapted ProMP conforms both to the boundary constraints and the restrictive original ProMP at the via-point.

\subsubsection{Smoothness constraints}
\label{sec:method_jconst_smooth}
Physical systems typically require smooth trajectories for operation and smooth trajectories also have the advantage of consuming less energy. 
For this reason, we consider imposing constraints or regularising the smoothness of a primitive.
Smoothness regularisation can be performed using the norm of the  $2^{\mathrm nd}$ time derivative of a primitive.
This is commonly referred to as spline smoothing regularisation \cite{williamsGaussianProcessesMachine2006} and it can be viewed as Bayesian estimation with a Gaussian process prior where the covariance function is determined by the $2^{\mathrm nd}$ order differential operator \cite{kimeldorfCorrespondenceBayesianEstimation1970}.
As a result, we can view the process $\{\bm{z}_t\}_{t \in [0,T]}$ as a Gaussian process with a covariance function, that is, a combination of the covariance functions corresponding to the given original ProMP and the smoothing regulariser. 

To simplify notation, we present the derivation for a scalar trajectory.
A smoothness regulariser can be formulated~as     
\begin{align*}
	R(\{z_t\})= \frac{1}{\lvert \mathcal{T} \rvert} \int_{\mathcal{T}} \vert z_t^{\prime\prime}\vert^{2} dt
\end{align*}
which in the context of ProMPs results in the quadratic~form 
\begin{align}
	\label{EqnSmoothQuad}
	R(\bm{w}) &= \bm{w}^{T}\left[\frac{1}{\lvert \mathcal{T} \rvert}  \int_{\mathcal{T}} [\phi_t^{\prime\prime}][\phi_t^{\prime\prime}]^{T} \!dt \right] \bm{w}.
\end{align}
We can either add $P_{\bm{w}}(R(\bm{w}) \leq d) \geq \alpha$ as a constraint to  \eqref{pf_const} or relax it to a regulariser $\kappa\, \E_{p(w)}[R(\bm{w})]$ to be added to the objective in \eqref{pf_kl2promp}. 

The distribution of the random variable $R(\bm{w})$ is a generalised-$\chi^2$ and due to positive semi-definiteness of the interaction matrix in \eqref{EqnSmoothQuad}, we have $R(\bm{w}) \geq 0$. Since the CDF has an intricate numerical form, we approximate $R(\bm{w})$ with a simpler Gamma distribution that matches its mean and variance, that is, we use $R(\bm{w}) \mathrel{\dot\sim} \Gamma(\alpha, \beta) $ with shape $\alpha = \E[R(\bm{w})] ^2/\operatorname{V}[R(\bm{w})] $ and rate $\beta = \E[R(\bm{w})] /\operatorname{V}[R(\bm{w})]$. By denoting with $\bm{\Phi}$ the matrix of the quadratic form in \eqref{EqnSmoothQuad}---which we compute by numerical integration---we find~that
 \begin{align}
 \E[R(\bm{w})] &= \bm{\mu}^{T}_{w}\bm{\Phi} \bm{\mu}_{w} + \mathrm{tr}(\bm{\Phi} \bm{\Sigma}_{w})
 \label{EqnSmoothessPenalty}
 \\
 \operatorname{V}[R(\bm{w})] &= 4\bm{\mu}^{T}_{w} \bm{\Phi} \bm{\Sigma}_{w} \bm{\Phi} \bm{\mu}_{w} + 2 \mathrm{tr}(\bm{\Phi}\bm{\Sigma}_{w}\bm{\Phi}\bm{\Sigma}_{w}).
 \nonumber
 \end{align}
 
 In Figure~\ref{fig:joint_smooth}, we illustrate the effect of a smoothness regulariser $\kappa\, \E_{p(w)}[R(\bm{w})]$  in a toy adaptation problem.
 The smoothness regularisation results in a trajectory distribution with a significantly reduced variance and a mean function that gets close to the constraint limits to increase smoothness.
 The significant reduction of variance  is due to the second term in \eqref{EqnSmoothessPenalty}. Denoting $\bm{\Phi} = \bm{U}_{\Phi}\diag(\lambda_{\phi}^i) \bm{U}_{\Phi}^{T}$ and writing $\bm{\Sigma}_{w} = \bm{U}_{\Phi} \bm{\Sigma}_{\Phi} \bm{U}_{\Phi}^{T}$, we find that $ \mathrm{tr}(\bm{\Phi} \bm{\Sigma}_{w}) = \sum_i \lambda_{\phi}^i \Sigma_{\Phi}^{ii}$. Since we have  $\mathrm{tr}(\bm{\Sigma}_{w}) = \mathrm{tr}(\bm{\Sigma}_{\Phi})  = \sum_i \Sigma_{\Phi}^{ii}$, we can conclude that \eqref{EqnSmoothessPenalty} incentivises small overall variances.
 While the KL objective in  \eqref{pf_kl2promp} is trying to keep both the mean $\bm{\mu}_w$ and covariance $\bm{\Sigma}_w$ close to the original ProMP's mean and covariance, the expression in  \eqref{EqnSmoothessPenalty} prefers $\bm{\mu}_{w}$ with (generally) smaller norms and covariances $\bm{\Sigma}_{w}$ with smaller overall variance. 
 As a result, by adding a smoothness regularisation, we express preference towards a small set of smooth trajectories within the ones defined by the original ProMP.
 As mentioned in Section \ref{SecBackgroundProMP}, it is generally hard to assess a-priori how much flexibility the ProMP's function class needs to be able to fit the task and yet not overfit either or both $p(\bm{w})$ and~$p_{0}(\bm{w})$.
 For this reason, choosing a flexible class and applying a smoothness regularisation is often the most convenient approach to take. 

\subsection{Task space constraints}
\label{sec:method_task}
\begin{figure*}[t]
  \centering
  \subfloat[Hyperplane]{
    \includegraphics[width=0.29\textwidth]{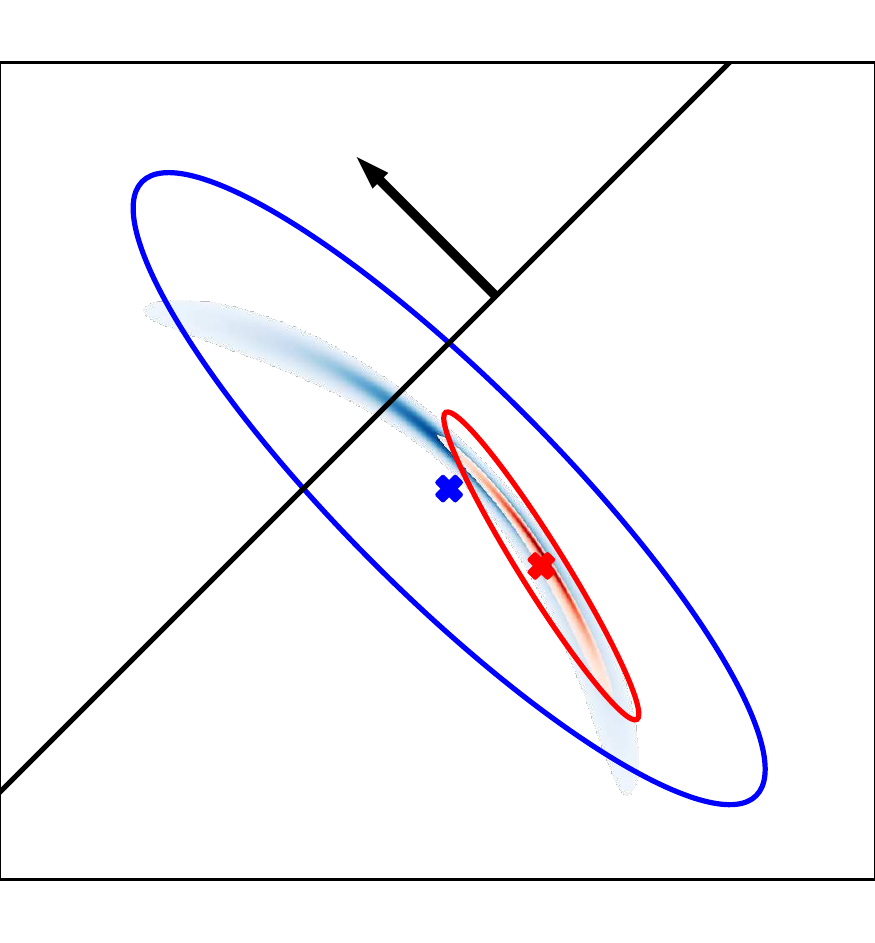}
    \label{sfig:hyperplane}
  }
  \hfil
  \subfloat[Waypoint]{
    \includegraphics[width=0.29\textwidth]{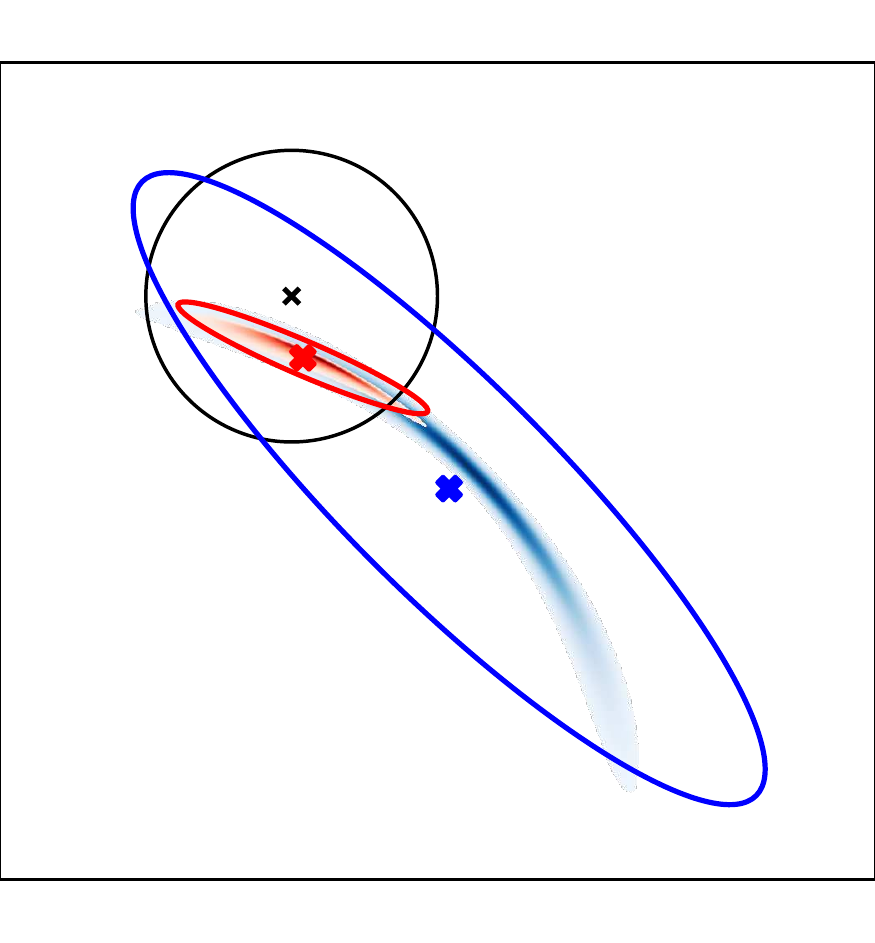}
    \label{sfig:waypoint}
  }
  \hfil
  \subfloat[Repeller]{
    \includegraphics[width=0.29\textwidth]{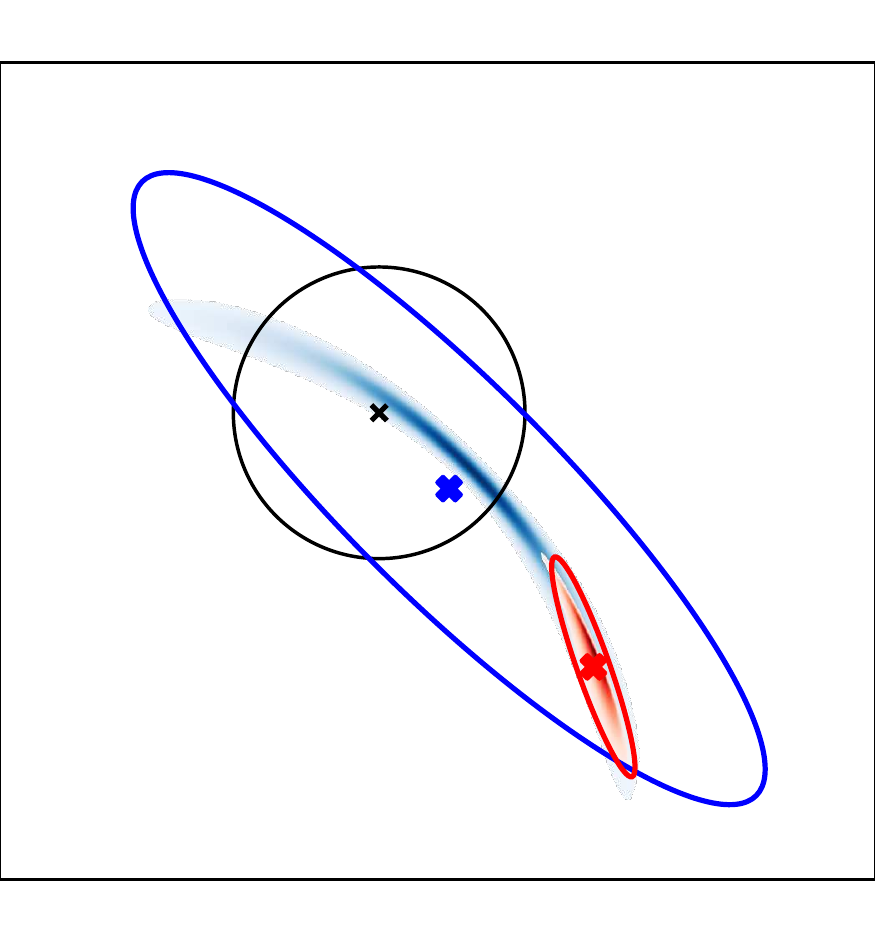}
    \label{sfig:repeller}
  }
  \caption{Approximations using the \emph{Unscented Transform}. The panels of the figure show how the \emph{unscented transform} based approximation works for a variety of constraints when using the forward kinematics function of a planar robot. The heatmaps show the original (blue) and the adapted (red) distribution of $\bm{x}_t(\bm{w})$ while the ellipsis show the Gaussians corresponding to their respective mean and covariance ($0.95$-level sets). }
  \label{fig:types_of_constraints}
\end{figure*}
It is equally important to formulate constraints on the Cartesian position of robot parts, like the end-effector, the elbow or any other point of interest determined by robot's joints' configuration.
Previous work used the probabilistic conditioning methods to adapt primitives in a table tennis setting with via-points \cite{gomez-gonzalezUsingProbabilisticMovement2016} and showed how obstacle avoidance can be achieved with policy search \cite{koertDemonstrationBasedTrajectory2016}, constrained optimisation \cite{colomeDemonstrationfreeContextualizedProbabilistic2017} and movement planning \cite{koertLearningIntentionAware2019}.
Our approach offers a generic way to incorporate limits on the robot's Cartesian coordinates by formulating constraints in terms of the process~$\{\bm{x}_t\}$.
Given a constraint $c_{k,t}(\bm{x}_t(\bm{w}))$, we approximate the distribution of the random variable $c_{k,t}(\bm{w}) = c_{k,t}(\bm{T}(\bm{z}_t(\bm{w})))$ with a distribution from a known family of distributions.
In cases where $\bm{T}$ or $c_{k}$ introduce nonlinear dependencies we use the \emph{Unscented Transform}~\cite{wanUnscentedKalmanFilter2000} to do moment-based approximations. When modelling in the Cartesian space, $\bm{T}$ is the identity function, hence the \emph{Unscented Transform} is not necessary.

The \emph{Unscented Transform} is a sample based method for computing statistics of a nonlinear transformation $\bm{g}(\bm{z}_{S})$ of the random variable $\bm{z}_{S}$, $S \subseteq [0, T]$.
Given $\bm{z}_{S}$ with mean $\E[ \bm{z}_{S} ]$ and variance $V[\bm{z}_{S}]$, one selects $2D+1$ specific sample points $\bar{\bm{z}}_m = \E[ \bm{z}_{S}] \pm (\alpha \sqrt{D V[\bm{z}_{S}]})_m$ and weights $W_m=1/(2(\alpha^2 D))$ to estimate the mean and the variance as
\begin{align*}
\E[\bm{g}(\bm{z}_{S})] &\approx \sum_m W_m \bm{g}(\bar{\bm{z}}_m) \\
V[\bm{g}(\bm{z}_{S})] &\approx \sum_m W_m \left[ \bm{g}(\bar{\bm{z}}_m) - \E[\bm{g}(\bm{z}_{S})]\right] \left[\bm{g}(\bar{\bm{z}}_m) - \E[\bm{g}(\bm{z}_{S})]\right]^T_{.}
\end{align*}
Here, the parameter $\alpha$ determines the spread of the sample points around the mean.
Depending on the type of constraint, we use the unscented transform either to approximate $\bm{x}_{S}(\bm{w})=\bm{T}(\bm{z}_{S}(\bm{w}))$ or $c_k(\bm{x}_{S}(\bm{w}))$ directly.

In the following we introduce various task space constraints.
We demonstrate most of these in small, dedicated experiments with a planar robot in Section~\ref{SecExpPlanar}, therefore we encourage readers to cross check the corresponding experiments for a visual interpretation of the individual effects.
\subsubsection{Hyperplane constraints}
\label{sec:method_task_hyperplane}
In a production setting robots are usually confined to their own workspace often delimited by fences.
We can formulate virtual wall constraints as
\begin{align}\label{tc_hyperplane}
  P_{\bm{w}}\big(\bm{n}_{k,t}^{T}(\bm{x}_t(\bm{w})-\bm{b}_{k,t}) \leq 0\big) \geq \alpha_{k,t},
\end{align}
where $\bm{n}_{k,t}$ and $\bm{b}_{k,t}$ denote the normal and bias vectors of a hyperplane. As a result, we have to estimate the distribution of the random variable $c_{k,t}(\bm{w}) = \bm{n}_{k,t}^{T}(\bm{x}_t(\bm{w})-\bm{b}_{k,t})$, which is a linear transformation of $\bm{x}_t(\bm{w})$. However, $\bm{x}_t(\bm{w})$ is non-linear in $\bm{w}$, thus we use the unscented transform to approximate $\bm{x}_t(\bm{w}) =\bm{ T}(\bm{z}_t(\bm{w}))$ with a Gaussian $\tilde{\bm{x}}_t(\bm{w}) \approx \bm{T}(\bm{z}_t(\bm{w}))$. This leads to the Gaussian approximation
\begin{align*}
c_{k,t}(\bm{w}) \mathrel{\dot\sim} \mathcal{N} \left(
	\bm{n}_{k,t}^{T}(\E[\tilde{\bm{x}}_t(\bm{w})]-\bm{b}_{k,t}), 
	\bm{n}_{k,t}^{T}\operatorname{V}[\tilde{\bm{x}}_t(\bm{w})] \bm{n}_{k,t} \right).
\end{align*} 
In this instance we set $d_{k,t} = 0$ and we use the corresponding Gaussian CDF in~(\ref{pf_loss}).

Note that although our approach allows for time-varying parameters, in most common settings both $\bm{n}_{k,t}$ and $\bm{b}_{k,t}$ are constant in time and  the constraints are required to be valid throughout the whole trajectory, that is, $\mathcal{T}_k=[0,T]$.
Using several \emph{hyperplane constraints} we can confine the trajectories to more complex convex domains.
We do this by requiring the satisfaction of the constraints for each individual hyperplane.

\subsubsection{Waypoints and repellers}
\label{sec:method_task_wp_rp}
In many applications we would like the trajectory of robot links or, more generally, any point of interest on the robot $\bm{x}_t(\bm{w})$, to be close to (waypoint) or to avoid (repeller) certain points $\bar{\bm{x}}_{t}$ in the task space. We formulate such trajectory constraints by
\begin{align}
  P_{\bm{w}} \big(\vert\bm{x}_t(\bm{w})-\bar{\bm{x}}_{t}\vert^2 \leq d^2\big) \geq \alpha_t
\end{align}
for waypoints and by
\begin{align}
  P_{\bm{w}} \big(\vert\bm{x}_t(\bm{w})-\bar{\bm{x}}_{t}\vert^2 > d^2\big) \geq \alpha_t
\end{align}
for repellent points. Depending on the choice of $\mathcal{T}$, these constraints can be used to implement end-point constraints $\mathcal{T}=\{T\}$, obstacle avoidance $\mathcal{T} = [0,T]$ or other constraints with more interesting temporal support $\mathcal{T} \subset [0,T]$.

To integrate these types of constraints into our framework we need to approximate the distribution of the random variable $c_{k,t}(\bm{w}) = \vert\bm{x}_t(\bm{w})-\bar{\bm{x}}_{t}\vert^2$. To approximate  $c_{k,t}(\bm{w})$ we use a Gamma approximation similar to the smoothness constraint in Section~\ref{sec:method_jconst_smooth}.
The mean $\E[c_{k,t}(\bm{w})]$ and the variance $\operatorname{V}[c_{k,t}(\bm{w})]$ are approximated using the unscented transform for~$c_{k,t}(\bm{w})$.
Depending on whether we have a waypoint or a repellent point we use the corresponding complementary values of the CDF of the resulting Gamma distribution.

Within the ProMP framework the adaptation with waypoints is commonly done by conditioning the primitive~\cite{paraschosProbabilisticMovementPrimitives2013,gomez-gonzalezUsingProbabilisticMovement2016}, which is fully compatible with our approach.
However, we would like to emphasise that our approach allows for a more expressive formulation of waypoints, as we can directly specify the physical margins as well as the temporal support of the waypoint.
When conditioning a primitive, the desired margins can only be achieved by tuning the variance of the waypoint and the temporal support is restricted to a single point in time.  

\subsubsection{Temporally unbound waypoints}
\label{sec:method_task_unboundWP}
In some applications one might not want to specify the point in time at which a via-point is reached.
Instead, one might only constrain the trajectory to visit the via-point in task space or joint space at an unspecified time point.
We give a practical example in Section~\ref{SecExpPandaDualHard}.
We can formalise this as 
\begin{align} \label{EqnBeenThere}
	 \max\limits_{t \in \mathcal{T}} P_{\bm{w}} \big(\vert\bm{x}_t(\bm{w})-\bar{\bm{x}}\vert^2 \leq d^2\big) \geq \alpha, \quad \mathcal{T} \subseteq [0,T]
\end{align}
that is, we require that there exists a time point $t_{\mathrm{max}} \in \mathcal{T}$ at which the waypoint constraint is satisfied with confidence~$\alpha$.
The l.h.s. of Equation \eqref{EqnBeenThere} is analytically intractable, however, its computation is algorithmically feasible since we can select $t_{\mathrm{max}} = \mathop{\mathrm{argmax}}_{t \in \mathcal{T}} P_{\bm{w}} \big(\vert\bm{x}_t(\bm{w})-\bar{\bm{x}}\vert^2 \leq d^2\big)$ when using the corresponding approximation for $P_{\bm{w}} \big(\vert\bm{x}_{t}(\bm{w})-\bar{\bm{x}}\vert^2 \leq d^2\big)$. Note that similarly to the \emph{smoothness constraint} this constraint is a \emph{path constraint} as the l.h.s. depends on the path~$\{\bm{x}_{t}\}_{t \in \mathcal{T}}$.

\subsubsection{Non-convex domains}
\label{sec:method_task_nonconvex}
With the hyperplane constraints in mind, we can also consider tackling the problem of obstacles with a piecewise linear shape, i.e., not spheres as in the case of repellers.
Specifically, we imagine the corner of a box that is to be avoided.
We can formulate this constraint as
\begin{equation}\label{eq:Pnonconvex}
P_{\bm{w}}\left(\bm{n}_{1,t}^{T}(\bm{x}_t-\bm{b}_{1,t}) \geq 0 \land \bm{n}_{2,t}^{T}(\bm{x}_t-\bm{b}_{2,t}) \geq 0\right) \leq 1 - \alpha_{t}.
\end{equation}
which can be expressed as the expectation constraint 
\begin{equation}\label{eq:Enonconvex}
\E_{p(\bm{w})}[H\big(\bm{n}_{1,t}^{T}(\bm{x}_t-\bm{b}_{1,t})\big) \cdot H\big(\bm{n}_{2,t}^{T}(\bm{x}_t-\bm{b}_{2,t})\big)] \leq 1 - \alpha_t.
\end{equation}
Computing this expectation is generally an analytically intractable problem because it requires computing the CDF of a multivariate Gaussian.
Therefore, we approximate the expectation of the product in \eqref{eq:Enonconvex} with a product of expectations. This results in approximating \eqref{eq:Pnonconvex} with
\begin{align*}
P_{\bm{w}}\left(\bm{n}_{1,t}^{T}(\bm{x}_t-\bm{b}_{1,t}) \geq 0\right) P_{\bm{w}}\left(\bm{n}_{2,t}^{T}(\bm{x}_t-\bm{b}_{2,t}) \geq 0\right) \leq 1 - \alpha_{t}.
\end{align*}
This approximation is exact when $\bm{n}_{1,t}^T\bm{n}_{2,t} = 0$, that is, the two hyperplanes are orthogonal to each other and either of the following conditions hold (i) $\operatorname{V}[\bm{x}_t]$ is diagonal or (ii) its eigenvectors are perpendicular to $\bm{n}_{1,t}$ and $\bm{n}_{2,t}$. 
The expression in equation \eqref{eq:Enonconvex} can be integrated in our framework by using the approach derived for approximating \eqref{tc_hyperplane}.

For arbitrary non-convex domains, however, one might have to derive approximations on a case by case basis.
A generic approach can be to find an optimal approximate spherical cover of the domain or its boundary, that is, a collection of spheres the union of which contain the domain, and use the corresponding collection of repellers for adaptation.
Another approach is to define a collection of time-dependent tangent hyperplane constraints with limited, say, a sliding window based, temporal support---in this case, however, the adaptation will be dependent on our choice for the constraints' temporal support or the constraints' parameters will depend on the robot's state.
Generally, a wide variety of domains (or their boundary) can be approximated by using an arrangement of repellers, hyperplanes and their temporal support.

\subsubsection{Mutual (self) avoidance}
\label{sec:method_task_dualAvoidance}
A common scenario where adaptation plays an important role is when several robots operate at the same time in a confined space.
Let us consider the simple scenario in which we would like two robots to execute independently learned ProMPs, however, we would also like them to avoid collision of their end-effectors or other points of interest.
We can formulate this simple adaptation problem as follows. Let $p_0(\bm{w}_1, \bm{w}_2) = p_{0}(\bm{w_1})\, p_0(\bm{w}_2)$ be the joint ProMP corresponding to the independently learned ProMPs of two robots and let us assume that we have two Cartesian points of interest $\bm{x}^{\scriptscriptstyle 1}_t(\bm{w}_1)$ and $\bm{x}^{\scriptscriptstyle  2}_t(\bm{w}_2)$, one on each robot, that should not collide. We can formulate this adaptation problem as learning a new joint ProMP $p(\bm{w})$ with  $\bm{w} = (\bm{w}_1, \bm{w}_2)$ such that 
\begin{align}\label{EqnJointCollision}
	P_{\bm{w}}( \vert \bm{x}^{1}_{t}(\bm{w}_1) -  \bm{x}^2_{t}(\bm{w}_2) \vert^2 > d^2  ) \geq \alpha_t.
\end{align}
We consider two options for the objective: (i) the Kullback-Leibler divergence $D_{\textrm{\sc KL}}[\,p(\bm{w}_1, \bm{w}_2) \vert\!\vert\, p_0(\bm{w}_1)\:p_0(\bm{w}_2)] $ between the jointly adapted ProMP and the joint distribution of the original ProMPs and (ii) the sum of marginal Kullback-Leibler divergences $D_{\textrm{\sc KL}}[\,p(\bm{w}_1) \vert\!\vert\, p_0(\bm{w}_1)]  + D_{\textrm{\sc KL}}[\,p(\bm{w}_2) \vert\!\vert\, p_0(\bm{w}_2)]$. The first objective favours adapted joint distributions that are similar to the factorising original distribution and thus penalises high covariance between $\bm{w}_1$ and $\bm{w}_2$---beyond what is necessary to satisfy the avoidance constraints, The second objective is more agnostic about the covariances and only focuses on the adapted marginal distributions being similar to the two original ProMPs.

In the planar robot toy experiment in Section~\ref{SecExpPlanarDual} we visualize the effect of the two objectives on the covariances.
This approach can be  generalised to several robots and several points of interest required in practical applications.
In  Section~\ref{SecExpPandaDualHard} we apply mutual avoidance with several points of interest in a real world dual arm setting.

\medskip\noindent
In this section we have presented a collection of constraints, and corresponding approximations, that enable us to apply the ProMP adaptation method we introduced in Section~\ref{SecMethProblem} to a large variety of adaptation problems. In the following we show how combining primitives could be achieved within our framework and we present the technical details of the Lagrangian optimisation.
Readers more interested in applications can skip to Section~\ref{sec:experiments}.

\subsection{Combining movement primitives}
\label{sec:method_combining}
Combining movement primitives is an important topic in the Learning from Demonstrations (LfD) literature. Although it is not the central topic of this paper, since we can always apply adaptation after solving the combination task, it is interesting to see how simple ProMP combinations could be integrated into our framework. 
There are several ways to combine ProMPs~\cite{paraschosProbabilisticPrioritizationMovement2017}, here we only consider  distributional geometric combinations.  We can combine and adapt $K$ ProMPs $p_i(\bm{w}),  \: i=1,\ldots, K$ in the joint space by choosing the objective in \eqref{pf_kl2promp}  as 
$\sum_i \alpha_i D[p(\bm{w}) \, \lvert\rvert \, p_i(\bm{w})]$, where $\alpha_i$ denotes a set of normalised weights.  When a ProMP is defined in the Cartesian space we can opt for marginal matching and replace the corresponding term with 
$(M/3T)\sum_t D[p(\bar{\bm{x}}_t(\bm{w})\, \lvert\rvert \,p_i(\bm{x}_t(\bm{w}))]$, where the scaling factor $M/3$ accounts for the difference in dimensionality. Here $\bar{\bm{x}}_t(\bm{w})$ is the approximation of Cartesian point of interest introduced above and $\bm{x}_t(\bm{w})$ are the Cartesian points of interest of the movement primitive to be combined.

\subsection{Optimisation techniques}
\label{SecOptimisation}

The approximation methods for the CDFs corresponding to the various constraints lead to an approximation of the Lagrangian \eqref{pf_loss} that we optimise using gradient methods.  For notational brevity, here we only present the optimisation for hyperplane, waypoint, and repeller constraints.
Let us use the notation  $\bm{\theta}_0 = (\bm{\mu}^0_{w}, \bm{\Sigma}^0_{w})$ for the parameters of the prior $p_0(\bm{w})$  and $\bm{\theta}_k = (\bm{n}_{k}, \bm{b}_{k})$  and $\bm{\theta}_k = (\bar{\bm{x}}, d)$ for the parameters of the hyperplanes, and repellent/waypoints, respectively.
These are fixed during the optimisation.
For the parameters of $p(\bm{w})$ we use $\bm{\theta} = (\bm{\mu}_{w}, \bm{\Sigma}_{w})$. With this notation, we can write the approximation of the  Lagrangian in \eqref{pf_loss} as
\begin{align}\label{EqnLagrangeAlgo}
\tilde{L}(\bm{\theta}, \{ \lambda_{k,t}\}) = D_{\textrm{\sc KL}}[\bm{\theta} \vert\!\vert\, \bm{\theta}_0] + \sum\limits_{k,t} \lambda_{k,t}\, \tilde{C}_{k,t}(\bm{\theta}, \bm{\theta}_k, \alpha_{k,t})
\end{align}
where we use $\tilde{C}_{k,t}(\bm{\theta}, \bm{\theta}_k, \alpha_{k,t}) =  \alpha_{k,t} - \tilde{F}_{c_{k,t}}(d_{k,t} ; \bm{\mu}_{w}, \bm{\Sigma}_{w}) $ to denote the corresponding constraint approximations. To solve the optimisation problem corresponding to \eqref{pf_dual},  we use an ascent-descent method which we detail in the following.

The term $D_{\textrm{\sc KL}}[\, \bm{\theta} \vert\!\vert\, \bm{\theta}_0] $ denotes the Kullback-Leibler divergence between two multivariate Gaussians
\begin{align}
\nonumber
D_{\textrm{\sc KL}}[\, \bm{\theta} & \vert\!\vert\, \bm{\theta}_0]  = 
	-\frac{1}{2}DM- \frac{1}{2} \log \vert \bm{\Sigma}_w \vert  
	+  \frac{1}{2}  \mathrm{tr}(\bm{[\Sigma}^{0}_{w}]^{-1}\bm{\Sigma}_{w})
	\\
	\nonumber
	&
	+ \frac{1}{2} (\bm{\mu}_{w}-\bm{\mu}^{0}_{w})^{T} \bm{[\Sigma}^{0}_{w}]^{-1} (\bm{\mu}_{w}-\bm{\mu}^{0}_{w}) + \frac{1}{2} \log \vert \bm{\Sigma}^{0}_{w} \vert 
\end{align}
which decouples into two independent terms for $\bm{\mu}_{w}$ and~$\bm{\Sigma}_{w}$ respectively. We use a Cholesky factorisation based parameterisation $\bm{\Sigma}_{w} = \bm{L}\bm{L}^{T},\bm{L} =  \bm{L}_{\mathrm{tril}} + \text{diag}(\exp (\bm{\gamma}))$, with $\bm{L}_{\mathrm{tril}}$ strictly lower triangular, to obtain $\frac{1}{2}\log \vert \bm{\Sigma}_{w} \vert  = \bm{1}^{T}\bm{\gamma}$ and thus to reduce 
$D_{\textrm{\sc KL}}[\bm{\theta} \vert\!\vert\, \bm{\theta}_0] $ to a simple analytic form. The factors $\tilde{C}_{k,t}(\bm{\theta}, \bm{\theta}_k, \alpha_{k,t})$ involve longer computation chains with numeric algebraic computations. We use automatic differentiation in \emph{Tensorflow} \cite{abadiTensorFlowLargeScaleMachine} to compute the gradients $\partial_{\bm{\theta}} \tilde{L}(\bm{\theta}, \{ \lambda_{k,t}\}) $ and perform gradient based descent steps.

The optimisation w.r.t. $\lambda_{k,t}$ requires~$\lambda_{k,t} \geq 0$. For this we use the Exponential Method of Multipliers (EMM) \cite{bertsekasNonlinearProgrammingSecond2003} to perform gradient quasi-ascent steps with  $\lambda_{k,t}^{\scriptscriptstyle (s+1)} = \lambda_{k,t}^{\scriptscriptstyle  (s)} \cdot \exp\{\eta_k \, \partial_{\lambda_{k,t} }\tilde{L}\}$; note that $ \partial_{\lambda_{k,t} }\tilde{L} = \tilde{C}_{k,t}$. The latter leads to $\Delta \lambda_{k,t}^{\scriptscriptstyle (s)} =  \lambda_{k,t}^{\scriptscriptstyle  (s)} \cdot (\exp\{\eta_k \, \partial_{\lambda_{k,t} }\tilde{L} \}-1)$ with $\Delta \lambda_{k,t}^{\scriptscriptstyle  (s)} \cdot \partial_{ \lambda_{k,t}} \tilde{L} \geq 0$, that is, the updates are aligned with the gradient. 

To optimise \eqref{EqnLagrangeAlgo} we use a double-loop algorithm: we perform several LBFGS \cite{nocedalNumericalOptimization2006} steps in 
$\bm{\theta}$ (\emph{inner-loop}), followed by a \emph{quasi-ascent} step in $\lambda_{k,t}$ with EMM (outer-loop). For the \emph{inner-loop} we tried several stopping criteria such as convergence of LBFGS and consistently increasing constraint violations. We observed that a combination of these two criteria leads to the best overall performance in terms of convergence speed.
The optimisation was implemented using the \emph{Tensorflow} automatic differentiation framework. The LBFGS  descent and EMM ascent optimisation is summarised in~Algorithm~\ref{alg:cpmp}.

\hl{The computational complexity of the optimisation algorithm scales with $D^3M^3$, and we have $DM + D^2M^2/2$ parameters to fit. Although developing an online algorithm is outside the scope of this paper, one can reduce the complexity of the algorithm by considering the following. To reduce the number of parameters one can use a low rank parameterisation $\bm{\Sigma}_w= \bm{U}\bm{U}^{T} + D$, where $\bm{U} \in \mathbb{R}^{DM\times K}$ and $\bm{D}$ is diagonal. By using the matrix determinant lemma, we have $\log \vert \bm{\Sigma}_w \vert = \log \vert \bm{I} + \bm{U}^{T}\bm{D}^{-1}\bm{U}\vert + \log \vert \bm{D} \vert $ and $\mathrm{tr}(\bm{[\Sigma}^{0}_{w}]^{-1}\bm{\Sigma}_{w}) = \mathrm{tr}(\bm{U}^{T}\bm{[\Sigma}^{0}_{w}]^{-1}\bm{U}) + \mathrm{tr}(\bm{[\Sigma}^{0}_{w}]^{-1}\bm{D})$ and thus we can reduce the computational complexity to $\max \{ K^3, K^{2}DM, K D^2 M^2\}$ and the number of parameters to fit to $(K+1)DM$. Sparse autoregressive structures for $\bm{\Sigma}^{-1}$ can also result in similar computational savings. Furthermore in an online setting one only has to infer the remaining part of the trajectory, this allows us to decrease $M$ as time progresses. Additionally, we can relax the non-imminent constraints, therefore, further reducing the computational cost.}

\begin{algorithm}[t]
  \begin{algorithmic}
    \State {\bf Inputs:} 
    \State $\bm{\theta}_0 = (\bm{\mu}^{0}_{w}, \bm{\Sigma}^{0}_{w})$ \Comment{original ProMP}
    \State $K$ constraints with $\theta_k$, $\alpha_{k,t}$, $\eta_k$, $T_k$, $\tau_k$
    \begin{itemize}
      \item Waypoint/Repeller: $\theta_k = (\bar{\bm{x}}_k, d_k)$ 
      \item Hyperplane: $\theta_k = (\bm{n}_{k}, \bm{b}_{k})$ 
    \end{itemize}
     \State {\bf Outputs:}
          \State $\bm{\mu}_{w}$, $\bm{\Sigma}^{1/2}_{w} =  \bm{L}_{\textrm{\sc tril}} + \text{diag}(\exp (\bm{\gamma}))$
    \State {\bf Learnable parameters:}
          \State $\bm{\theta} = (\bm{\mu}_{w}, \bm{L}_{\textrm{\sc tril}}, \bm{\gamma})$, $\{\lambda_{k,t}\}$
    \State {\bf Initialisation:} 
    	\State $\bm{\theta}^{0} \gets \bm{\theta}_0, \: \lambda_{k,t}^{0} \:  (\text{see Section~\ref{sec:experiments}})$
     \State {\bf Opimisation:}
    \Repeat
      \Repeat \Comment{Descent} 
      \State \bm{$\theta}^{(s+1)} \gets \text{LBFGS-Step}(\tilde{L}, \partial_{\theta}\tilde{L}, \bm{\theta}^{s}, \{ \lambda_{k,t}^{(r)}\})$                 
       \Until{inner-loop-condition}
      \State $\lambda_{k,t}^{(r+1)} \gets \lambda_{k,t}^{(r)} \cdot \exp\{\eta_k \, \tilde{C}_{k,t}(\bm{\theta}^{s}, \bm{\theta}_k, \alpha_{k,t})\}$ \Comment{Ascent}
    \Until{converged}
  \end{algorithmic}
    \caption{A gradient ascent-descent optimisation algorithm (LBFGS-EMM) to optimise \eqref{pf_dual} with the objective \eqref{EqnLagrangeAlgo}.}
    \label{alg:cpmp}
\end{algorithm}
\clearpage{}%

\clearpage{}%
\section{Experimental Results}
\label{sec:experiments}

We validate our approach in two different settings.
First, we demonstrate how our proposed adaptation method works through experiments for all constraint types introduced in Section~\ref{sec:method_task} on a simulated, 4-DoF planar robot arm.
We then advance to quantitative evaluations of our method including a comparison to DEBATO~\cite{koertDemonstrationBasedTrajectory2016}, another state-of-the-art approach for primitive adaptation. 
Finally, we place two Franka Emika Panda robot arms with 7~DoF in the same workspace and we design multiple experiments, each of which require several different types of constraints.

\subsection{Experiments on a simulated planar robot arm}
\label{SecExpPlanar}

\begin{figure*}[t]%
  \begin{tabular}{lccr}
    \resizebox*{0.23\textwidth}{!}{%
      \includegraphics{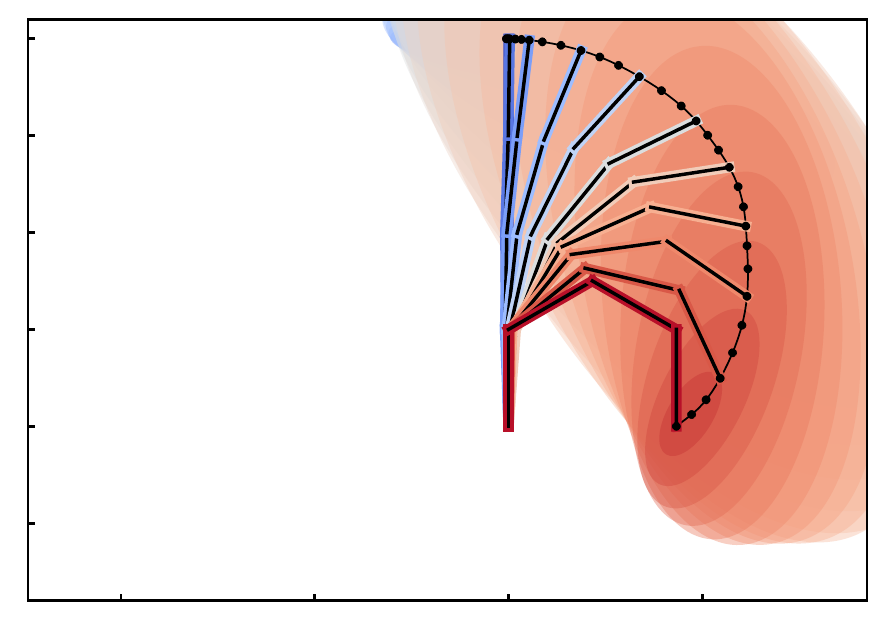}
      \put(-270,70){\rotatebox{90}{{\Large Original}}}
    }&%
    \multicolumn{2}{c}{
      \resizebox*{0.46\textwidth}{!}{
        \includegraphics{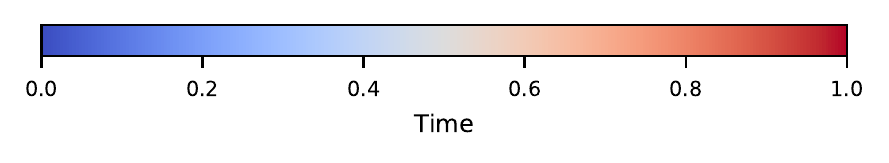}
      }%
    }&%
    \resizebox*{0.23\textwidth}{!}{%
      \includegraphics{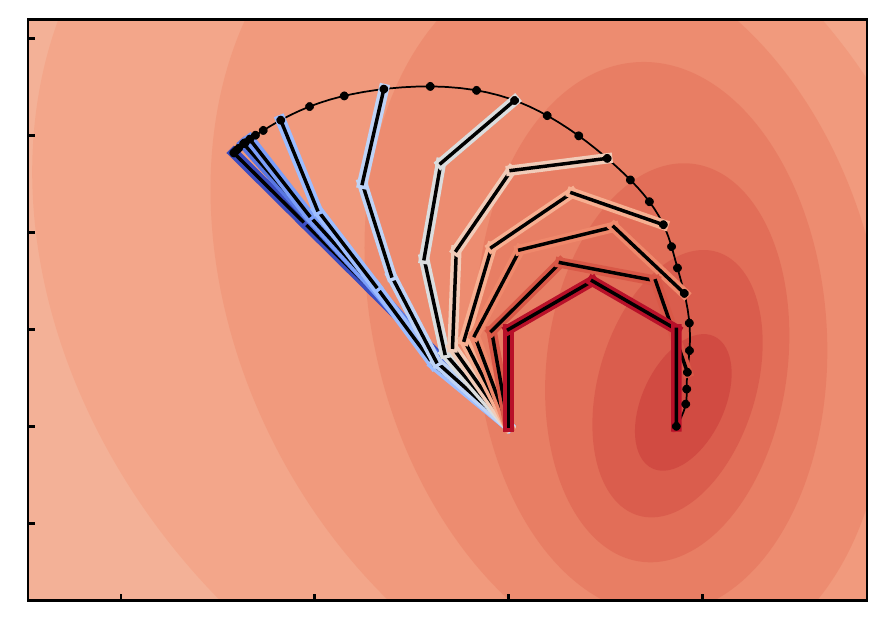}
    }\\
    \resizebox*{0.23\textwidth}{!}{%
      \includegraphics{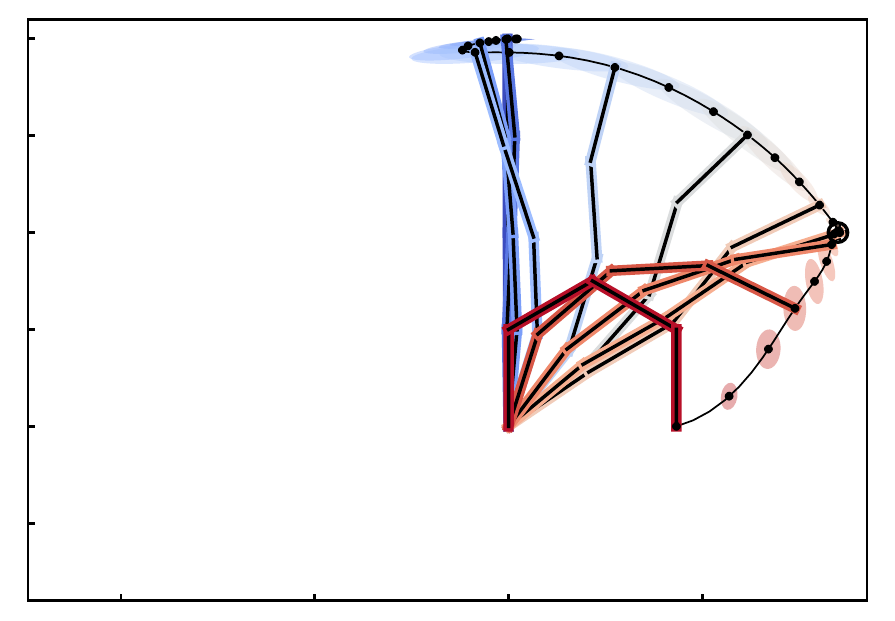}
      \put(-270,70){\rotatebox{90}{{\Large only KL}}}
      \put(-150,180){{\Large waypoint}}
    }&%
    \resizebox*{0.23\textwidth}{!}{%
      \includegraphics{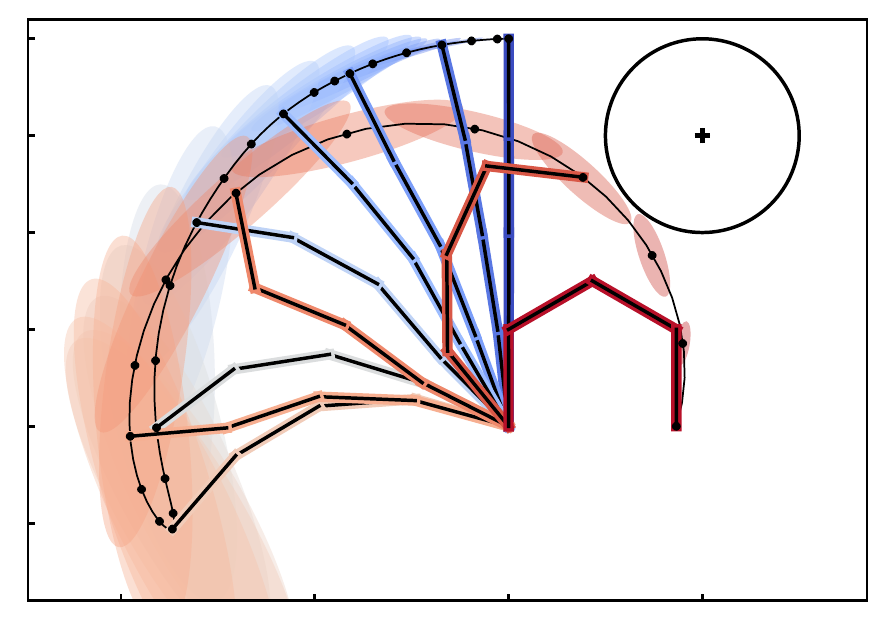}
      \put(-150,180){{\Large repeller}}
    }&%
    \resizebox*{0.23\textwidth}{!}{%
      \includegraphics{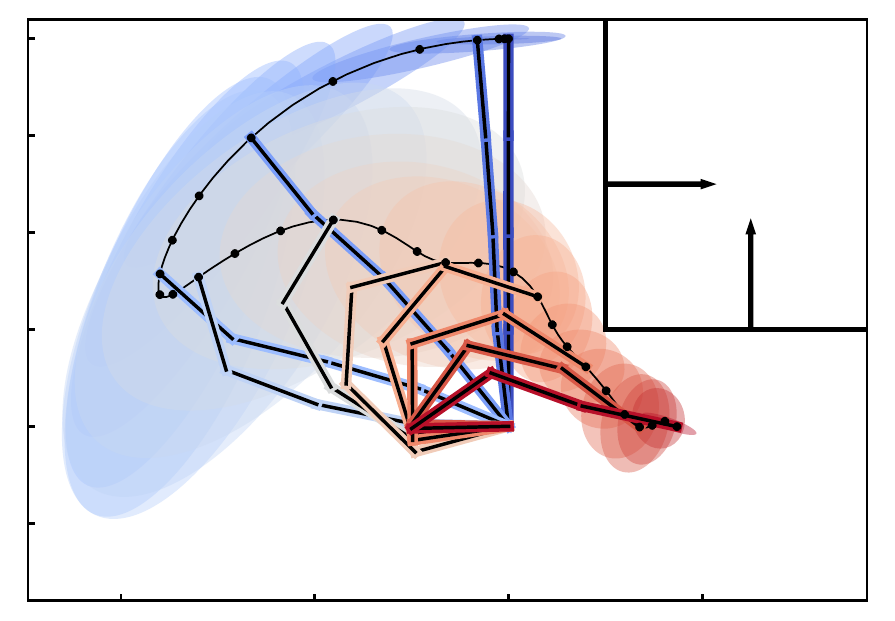}
      \put(-150,180){{\Large non-convex}}
    }&%
    \resizebox*{0.23\textwidth}{!}{%
      \includegraphics{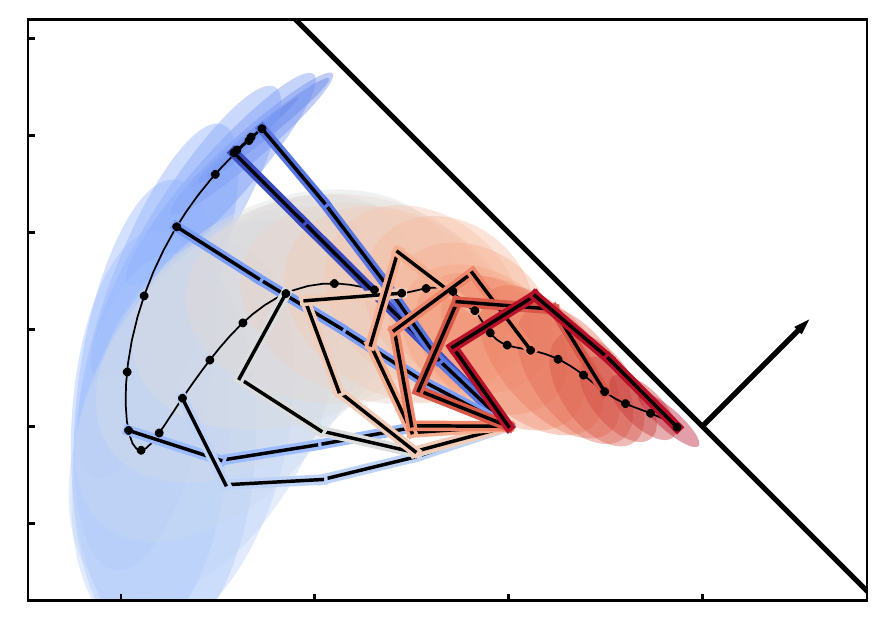}
      \put(-150,180){{\Large hyperplane}}
    }\\
    \resizebox*{0.23\textwidth}{!}{%
      \includegraphics{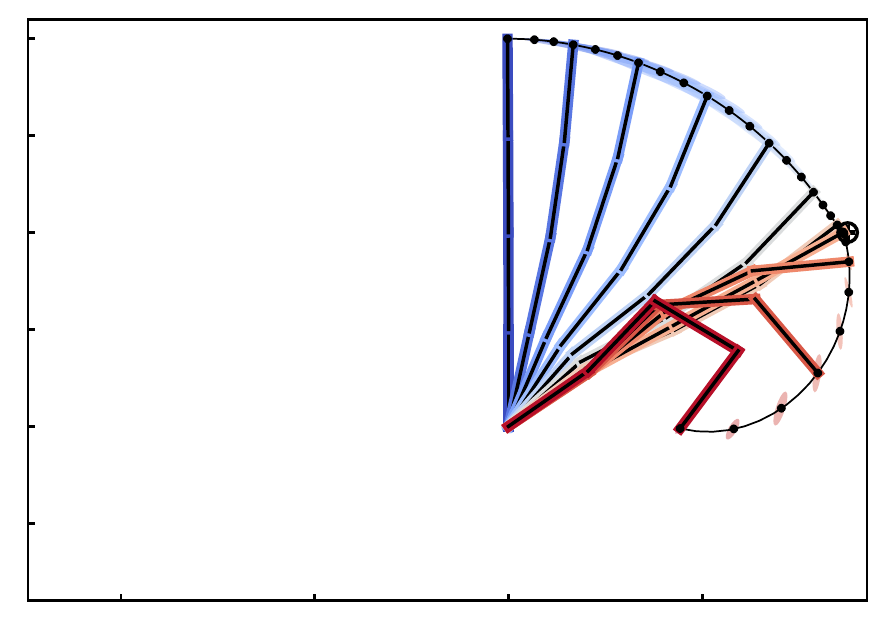}
      \put(-270,30){\rotatebox{90}{{\Large KL with smoothness}}}
    }&%
    \resizebox*{0.23\textwidth}{!}{%
      \includegraphics{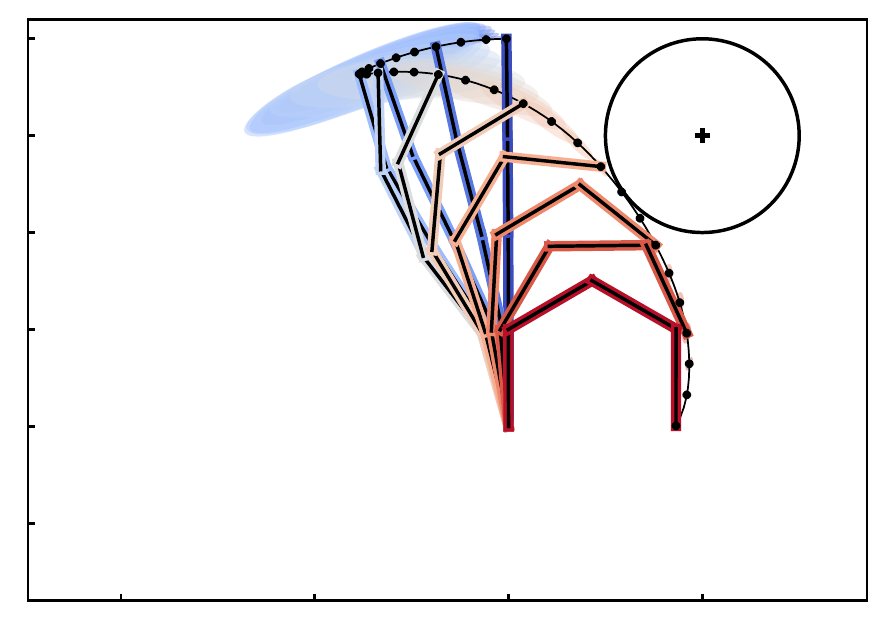}
    }&%
    \resizebox*{0.23\textwidth}{!}{%
      \includegraphics{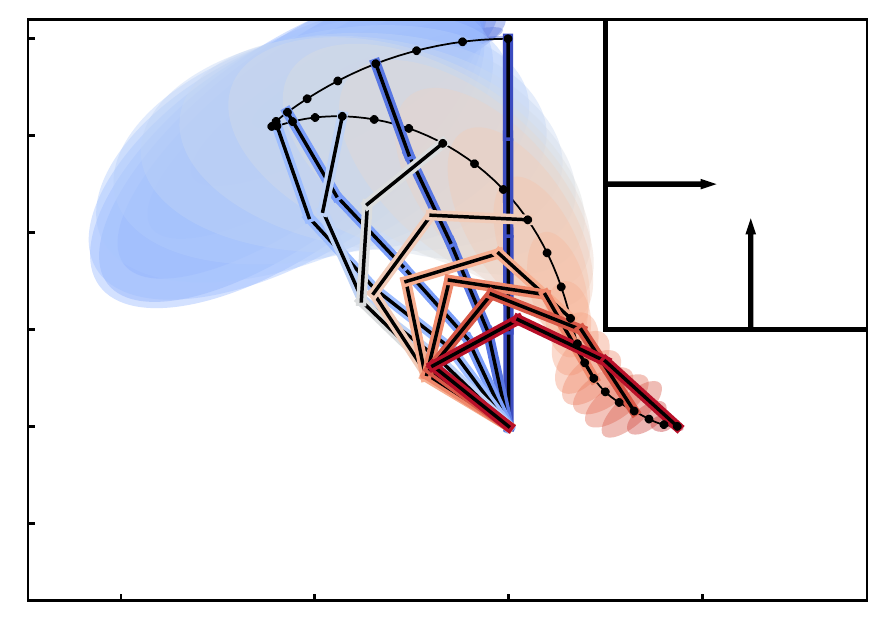}%
    }&%
    \resizebox*{0.23\textwidth}{!}{%
      \includegraphics{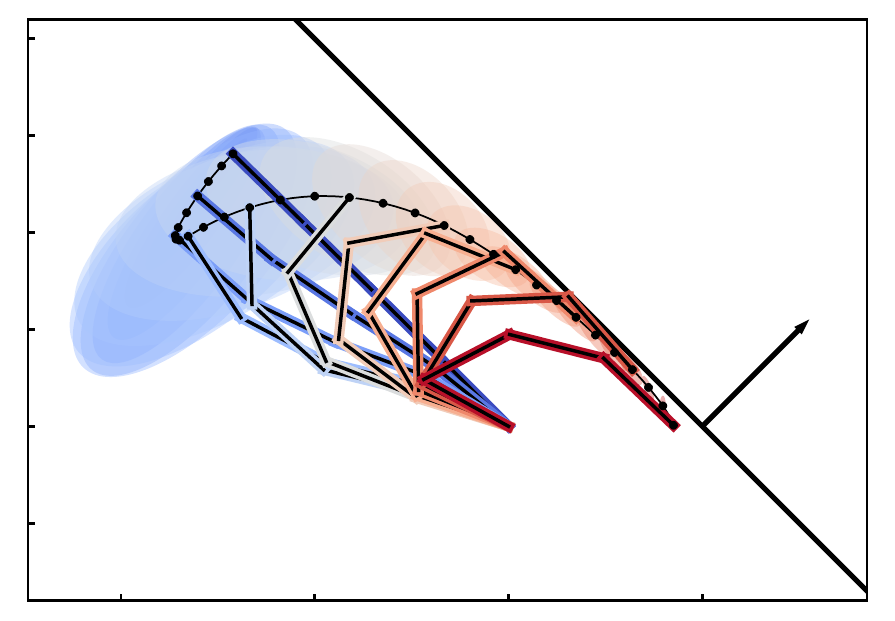}
    }
  \end{tabular}
  \caption{
    \emph{Illustrating trajectory adaptation on a planar robot.}  The panels in this figure illustrate the examples presented in Section~\ref{SecExpPlanar}.
    The colour gradients show time progression.
    We visualise the mean trajectory and the 0.95 level sets corresponding to the end-effector's Cartesian space covariances computed using the unscented transform.
    Note that the covariances are Gaussian approximations to the true distribution of the end-effector position and thus can also show mass outside of the robot's reach (see also Figure~\ref{fig:types_of_constraints}).
    The top row shows the original, unconstrained ProMPs after learning from demonstrations.
    For the experiments with the waypoint, the repeller and the non-convex constraint these are the same (top-left), whereas the original ProMP for the hyperplane experiment is different and is visualised in the top right panel.  
    The rest of the panels show adaptation for waypoint (left column),  repeller (second from left), non-convex (second from right), and hyperplane (right column) constraints; both without (middle row) and with (bottom row) smoothness regularisation added to the objective.
    See Section~\ref{SecExpPlanar} for details.
  }
  \label{FigExpPlanar}
\end{figure*}

In order to demonstrate each task space constraint, we implemented a simple planar robot arm simulator with linear dynamics.
The system state consists of the joint positions and angular velocities while the angular accelerations are directly used as control input.
We use the ProMPs to model the joint positions.
We use the mean and variance of the Cartesian coordinates corresponding to the adapted ProMP $\mathcal{N}(\bm{w}; \bm{\mu}_{w}, \bm{\Sigma}_{w})$ to create the illustrations.
Sampling trajectories to learn the original (unadapted) ProMP, i.e., $\mathcal{N}(\bm{w}; \bm{\mu}^{0}_{w}, \bm{\Sigma}^{0}_{w})$ is carried out by: (i) adding an attractor for the end position to the linear system, (ii) fixing the starting position, (iii) i.i.d. sampling accelerations to generate trajectories (iv) learning the original ProMP from the sampled trajectories, and finally, (v) conditioning the ProMP on the final joint configuration.
The trajectory distributions corresponding to the resulting ProMPs are shown in the top panels of Figure~\ref{FigExpPlanar}.
We choose high variance original ProMPs in order to better demonstrate the effect of the constraints. 
The experiments in Section~\ref{SecExpPandaDualSimple}, however, demonstrate a more balanced interplay between the original ProMPs and the constraints.

In all adaptation problems presented in the following we choose as objective the Kullback-Leibler (KL) divergence either with or without the smoothness regularisation term, as presented in Section~\ref{sec:method_jconst_smooth}.
The smoothness regularisation term consists of a weighted sum of independent regularisations per each joint with joints closer to the base having a higher regularisation weight.
For most tasks we used a weighting factor of 0.1 for the overall smoothness regularisation term, while within this term, we used a weighting of $[2.0, 0.1, 0.1, 0.1]$ for the corresponding 4 joints.
This weighting reduces the movement of the first link, which has to move the greatest mass and thus is also the most energy consuming.

We add narrow joint limit constraints for the starting joint angles ($t=0$) to fix the initial position.
In addition to this start-point constraint, we add a waypoint constraint for the end-effector's Cartesian position $\bm{x}^{\mathrm{end}}_{t}(\bm{w})$ at the end of the time interval ($t=T$) to fix the final position the robot has to reach in Cartesian space.
In the following we refer to the end-point of the $4^{\mathrm{th}}$ link of the robot as the end-effector.
We use location $\tilde{\bm{x}}= (1.73, 0.0)$ with a radius of $d=0.05$ for the waypoint constraint at the final time-point.
For all constraints in the experiments we choose a confidence level of $\alpha=0.999$.
Each arm link has length 1.0.

All optimisations are carried out using Algorithm~\ref{alg:cpmp}.
The initial value for $\bm{\theta} =(\bm{\mu}_{w}, \bm{\Sigma}_{w})$ is set to $\bm{\theta}^{s=0}=(\bm{\mu}^{0}_{w}, \bm{\Sigma}^{0}_{w})$.
The initial values for the Lagrange multipliers $\lambda_{k,t}$ corresponding to the start time (initial joint configuration) and end time (end-effector's Cartesian position) are set to $\lambda_{k,t}^{r=0}=100$ while the ones corresponding to the rest of the constraint are set to $\lambda_{k,t}^{r=0}=1.0$.

\subsubsection{Waypoint constraints}
\label{SecExpPlanarWaypoint}
We demonstrate the waypoint constraint by setting a waypoint at location $\bm{\bar{x}}=(3.4, 2.0)$ with radius $d=0.05$ and force the end-effector $\bm{x}^{\mathrm{end}}_{t}(\bm{w})$ to stay at that position for one-tenth of the trajectory.
Therefore the corresponding constraint is chosen to be active during the time interval $\mathcal{T}=[0.65,0.75] \subset [0,1]$.
The left-column panels in Figure~\ref{FigExpPlanar} show the resulting motions' means and covariances, with 0.95 level sets, for $\bm{x}^{\mathrm{end}}_{t}(\bm{w})$ for the KL-only (middle row) and the KL with smoothness regularised (bottom row) objective, respectively.

When using the KL objective, both the final configuration and generally the overall motion pattern resembles the original ProMP.
In contrast when adding the smoothness penalty we obtain a trajectory, which shows clear deviations from the original ProMP.
Especially the last part of the trajectory is changed such that the first link moves as little as possible, a behaviour which is incentivised by the smoothness regularisation.
Nonetheless the movement stays within the limits created by the original ProMP's confidence bounds.

\subsubsection{Repeller constraints}
\label{SecExpPlanarRepeller}
In this experiment we place a repeller with radius $d=1.0$ at location $\bm{\bar{x}}=(2.0, 3.0)$.
The end-effector $\bm{x}^{\mathrm{end}}_{t}(\bm{w})$ is required to satisfy the repeller constraint during the whole length of the trajectory, hence $\mathcal{T}=[0,1]$.
The second-column panels in Figure~\ref{FigExpPlanar} show the resulting mean trajectory and the covariances ($0.95$ level sets) for $\bm{x}^{\mathrm{end}}_{t}(\bm{w})$ for the KL-only (middle row) and the smoothness regularised (bottom row) objective, respectively.
 
The trajectory adaptation resulting from the  KL objective can be explained as follows:
First, moving the first joint has a large effect on the position of the robot relative to the repelling point, which allows the robot to keep the changes for the rest of the joints minimal compared to the original ProMP.  Second, moving further away from the repeller gives the trajectory the possibility to keep a higher variance for the most part of the trajectory.
Therefore the KL objective paired with a high variance original ProMP leads to such solutions being favourable.
Adding a smoothness regulariser with a proportionally larger weight on the first joint (bottom row) results in a seemingly more natural movement.
The latter choice of objective is also plausible in a real environment as the first joint moves more weight and thus consumes more energy.

\subsubsection{Non-convex domain constraints}
\label{SecExpPlanarNonConvex}
We place a corner defined by two hyperplanes $\bm{n}_1 = (1,0), \bm{b}_1=(0.5, 0.5)$ and $\bm{n}_2 = (0,1), \bm{b}_2=(0.5, 0.5)$.
In this case, we require not only the end-effector $\bm{x}^{\mathrm{end}}_{t}(\bm{w})$ to satisfy the constraint, but we also impose the same limitations on the ends of the second and the third link, $\bm{x}^{\mathrm{2}}_{t}(\bm{w})$ and $\bm{x}^{\mathrm{3}}_{t}(\bm{w})$.
The second column from the right in Figure~\ref{FigExpPlanar} shows the resulting means and covariances ($0.95$ level sets) for $\bm{x}^{\mathrm{end}}_{t}(\bm{w})$ for the KL-only (middle row) and the smoothness regularised (bottom row) objective, respectively.

As expected, the resulting trajectory shows a similar behaviour as in  the case of the repeller, however, 
the configuration of the robot and the non-convex constraint do not allow the robot to keep the final configuration given by the prior.
Instead it has to resort to a different motion at this part of the trajectory.
Similarly to the repeller experiment, the KL-only objective favours a  motion with higher variance, in which the robot can also keep at least part of the trajectory for the last three joints similar to the original ProMP.

\subsubsection{Hyperplane constraints} 
\label{SecExpPlanarHyperplane}
To demonstrate how a hyperplane constraint affects the motion, we choose a hyperplane with $\bm{n} = (1.0,1.0)$, $\bm{b}=(0, 2)$ and we require the end-effector $\bm{x}^{\mathrm{end}}_{t}(\bm{w})$ and the penultimate joint $\bm{x}^{3}_{t}(\bm{w})$ to satisfy it.
Similary to the experiment with non-convex domains, in this setup the movement and the final configuration have to be significantly different to that of the original ProMP.
The right-column panels in Figure~\ref{FigExpPlanar} show the resulting means and covariances, with $0.95$ level sets, for $\bm{x}^{\mathrm{end}}_{t}(\bm{w})$ for the KL-only objective (middle row) and the weighted sum with smoothness regularisation (bottom row), respectively. Note that the original ProMP is depicted in the top right panel, where the original movement mainly has a different start configuration and a higher variance.

We observe that the adapted primitive results in a movement similar to the experiments with the repeller and the non-convex constraint.
The robot first moves away from the constraint boundary in order to take a configuration that makes the final objective easier to reach.
As we have seen in the two previous cases, the KL-only objective produces a high variance trajectory, whereas the smoothness regularisation leads to a reduced variance and more a direct approach towards the target end-point.
Note that modelling the primitive in joint space allows us to force not only the end-effector, but also the whole robot arm to stay inside the domain defined by the hyperplane.

\subsubsection{Mutual avoidance constraints}
\label{SecExpPlanarDual}
\begin{figure}
    \centering
    \subfloat[Original]{
        \includegraphics[width=0.45\columnwidth]{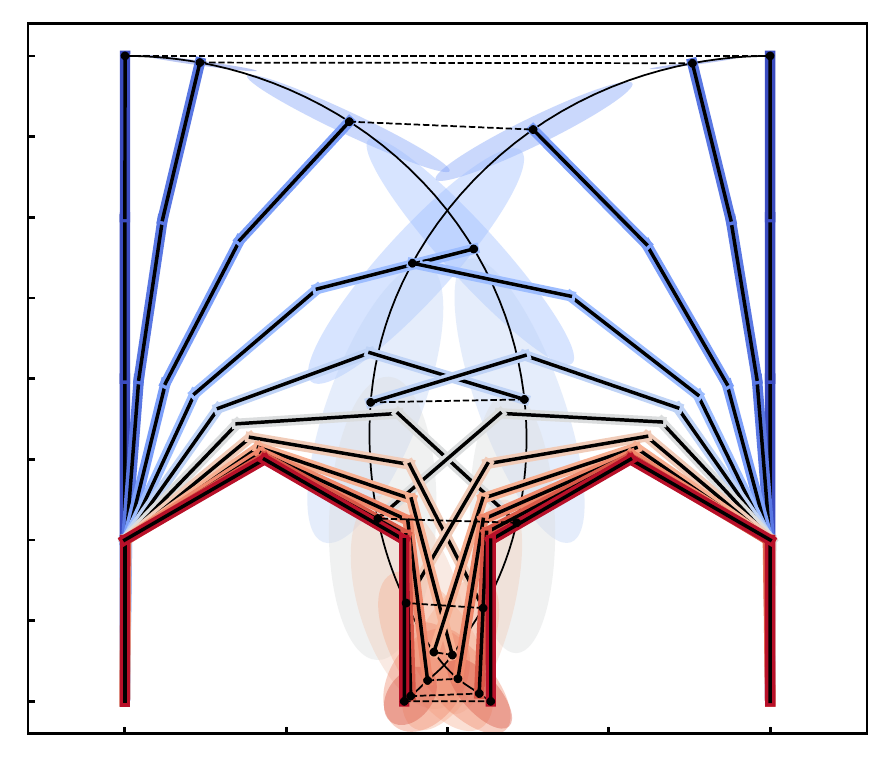}
        \label{sfig:planar_dualR_prior}
    }
    \hfil
    \subfloat[Adapted]{
        \includegraphics[width=0.45\columnwidth]{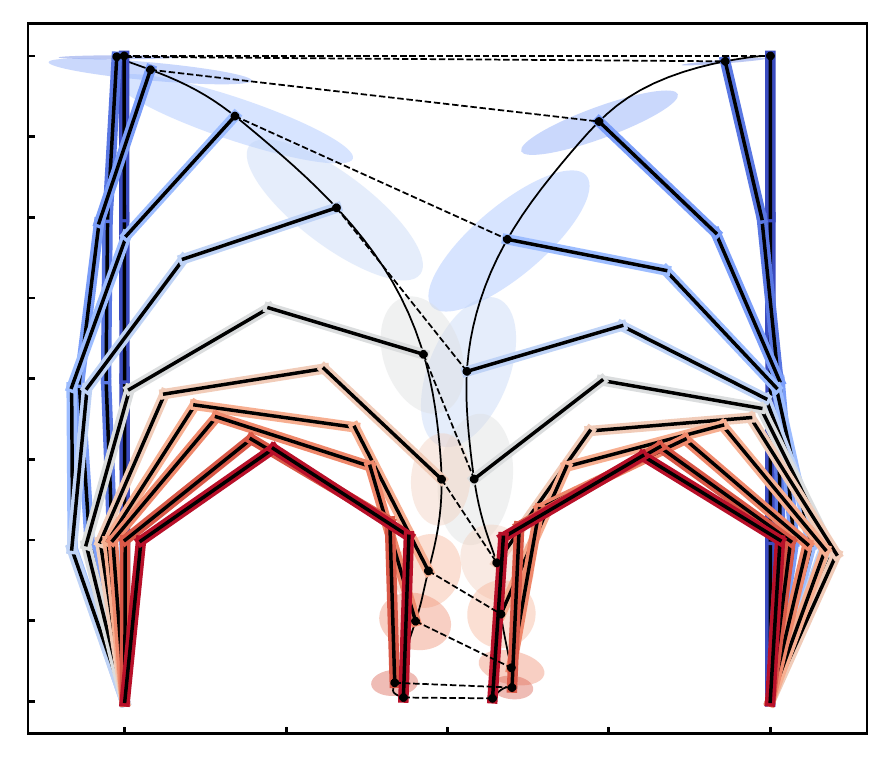}
        \label{sfig:planar_dualR_posterior}
    }
    \caption{\emph{Illustrating mutual avoidance for two planar robots.}  In this figure we show the results of the mutual avoidance experiment in Section~\ref{SecExpPlanarDual}.
    The left panel shows the independently learned original ProMPs, the right panel shows the combined ProMP, adapted with the mutual avoidance constraint from Section~~\ref{sec:method_task_dualAvoidance}.
    Analogous to Figure~\ref{FigExpPlanar}, the colour gradients show time progression and the shaded ellipses stand for the end-point's covariances.
    Mutual avoidance is realised by the left robot slowing down and the right robot speeding up its motion.}
    \label{fig:planar_dualR}
\end{figure}
\begin{figure}
    \centering
    \includegraphics[width=0.98\columnwidth]{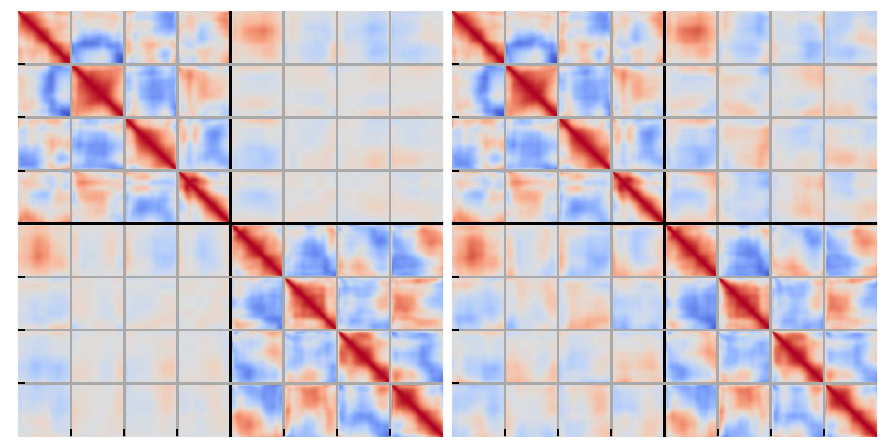}
    \caption{{\em Illustrating the choice of objective for the mutual avoidance problem.}
    The panels show the correlations in the adapted joint trajectories when using the objectives $D_{\textrm{\sc KL}}[\,p(\bm{w}_1, \bm{w}_2) \vert\!\vert\, p_0(\bm{w}_1)\:p_0(\bm{w}_2)]$ (left panel) and $D_{\textrm{\sc KL}}[\,p(\bm{w}_1) \vert\!\vert\, p_0(\bm{w}_1)]  + D_{\textrm{\sc KL}}[\,p(\bm{w}_2) \vert\!\vert\, p_0(\bm{w}_2)]$ (right panel).
    The colour indicates the correlation from -1 (blue) through 0 (grey) to +1 (red).
    The small blocks stand for the robot's joints, whereas the large ones stand for the two robots.
    As expected, the covariance agnostic objective leads to higher correlations between the robots, as can be seen when comparing the lower left and the upper right quadrants.
    }
    \label{fig:planar_dualR_correlations}
\end{figure}
Besides illustrating the mutual avoidance constraint, in this experiment we investigate the effect of the two different options for choosing the KL objective, as described in Section~\ref{sec:method_task_dualAvoidance}.
We designed a dual-arm setting with two planar robot arms, which are placed in the configurations shown in Figure~\ref{fig:planar_dualR}.
This configuration would lead to a collision should the trajectories be executed simultaneously.
The corresponding original ProMPs are learnt independently.
We set one mutual avoidance constraint for the end-effectors, namely, $P_{\bm{w}}( \vert \bm{x}^{\mathrm{\scriptscriptstyle1, end}}_{t}(\bm{w}) -  \bm{x}^{\mathrm{\scriptscriptstyle 2, end}}_{t}(\bm{w}) \vert^2  >  d_{\mathrm{end}}^2  ) \geq \alpha$ and two constraints for the end-effectors and the last joints  $P_{\bm{w}}( \vert \bm{x}^{\mathrm{ \scriptscriptstyle 1, 4}}_{t}(\bm{w}) -  \bm{x}^{\mathrm{ \scriptscriptstyle 2, end}}_{t}(\bm{w}) \vert^2 > d_{\mathrm{c}}^2  ) \geq \alpha$   and $P_{\bm{w}}( \vert \bm{x}^{\mathrm{ \scriptscriptstyle 1, end}}_{t}(\bm{w}) -  \bm{x}^{\mathrm{ \scriptscriptstyle 2, 4}}_{t}(\bm{w}) \vert^2 > d_{\mathrm{c}}^2  ) \geq \alpha$ , respectively.
We use $ d_{\mathrm{end}} = 0.4$ and $ d_{\mathrm{c}}=0.8$ for the corresponding distances.
We then adapt an extended joint model with weights $\bm{w} = (\bm{w}_1, \bm{w}_2)$, see Section~\ref{sec:method_task} for details.
The original ProMP and the adapted ProMP are shown in Figure~\ref{fig:planar_dualR}.

We observe that the collision is avoided by a speed-up of the right-robot and a slow-down of the left-robot's motion, when compared to the original ProMPs.
The motion pattern and the final configurations, however, remain similar.
The optimisation problem has two modes, either the left or the right robot speeds up its motion. 
Depending on the trajectories sampled to learn the two original ProMPs, the optimisation chooses one of the two modes.
Indeed, we observed a correlation between small fluctuations in the original ProMPs and the choice of mode in the optimisation.

When comparing the two objectives we propose in Section~\ref{sec:method_task}, we were particularly interested in the effect they have on the learnt correlations between the motion of the two robots.
In Figure~\ref{fig:planar_dualR_correlations}, we show the correlation corresponding to the joint space coordinates of the two robot arms.
While both objectives result in trajectory samples with no collisions, $D_{\textrm{\sc KL}}[\,p(\bm{w}_1, \bm{w}_2) \vert\!\vert\, p_0(\bm{w}_1)\:p_0(\bm{w}_2)]$ (left panel) tends to penalise high correlation.
Note that the learned correlations between links or between robots can encode useful task specific information, as is shown in Section~\ref{sec:exp_dualCross_correl}.

In this Section we carried out a series of experiments to show how our trajectory adaptation method works in a simulated environment.
We designed tasks that shed light on how each constraint introduced in Section~\ref{sec:method_task} can be used and showed how the different weightings of the Kullback-Leibler divergence---similarity to the original ProMP---and the smoothness regularisation term affect the resulting trajectories.

\subsection{Quantitative evaluation}

In this Section we quantitatively evaluate three of our task space constraints: the repeller, the temporally unbound waypoint and the hyperplane constraint on randomly generated two-dimensional environments.
These three constraints are the basis for all the task-space constraints in our framework, that is, the mutual avoidance constraint is implemented as a moving repeller and the non-convex constraints are non-convex combinations of multiple hyperplane constraints.
For the repeller constraint we compare our approach to prior work, for the adaptation with temporally unbound waypoints and hyperplanes there is no prior work to compare to.

\subsubsection*{Comparison to DEBATO~\cite{koertDemonstrationBasedTrajectory2016}}
\label{sec:exp_comp_debato}
\begin{table*}
  \centering%
  \caption{Comparison of our approach to DEBATO~\cite{koertDemonstrationBasedTrajectory2016} on randomly generated two-dimensional obstacle avoidance problems.}
    \begin{tabular}{cccccccc}
    \toprule
    \# of obstacles & n & failed (ours) & failed (DEBATO) & Obstacle violations (ours) & Obstacle violations (DEBATO) & KL (ours) & KL (DEBATO) \\
    \midrule
    1 & 1000 & 0 (0.0\%) & 79 (7.9\%) & 0.30\% $\pm$ 0.09\% & 0.21\% $\pm$ 0.98\% & 0.28 $\pm$ 0.08 & 1.61 $\pm$ 1.02 \\
    2 & 1000 & 7 (0.7\%) & 173 (17.3\%) & 0.41\% $\pm$ 0.15\% & 0.21\% $\pm$ 1.14\% & 0.47 $\pm$ 0.14 & 2.19 $\pm$ 1.14 \\
    3 & 1000 & 11 (1.1\%) & 221 (22.1\%) & 0.48\% $\pm$ 0.17\% & 0.33\% $\pm$ 1.91\% & 0.60 $\pm$ 0.18 & 2.67 $\pm$ 1.46 \\
    \bottomrule
    \end{tabular}
  \label{tab:exp_comp_debato}%
\end{table*}

\begin{figure}
    \centering
    \includegraphics[width=0.98\columnwidth]{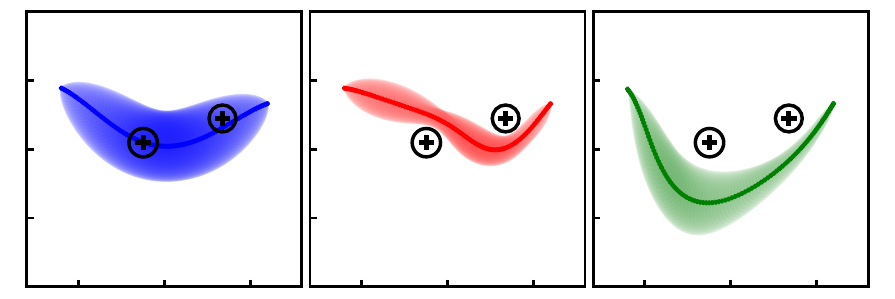}
    \includegraphics[width=0.98\columnwidth]{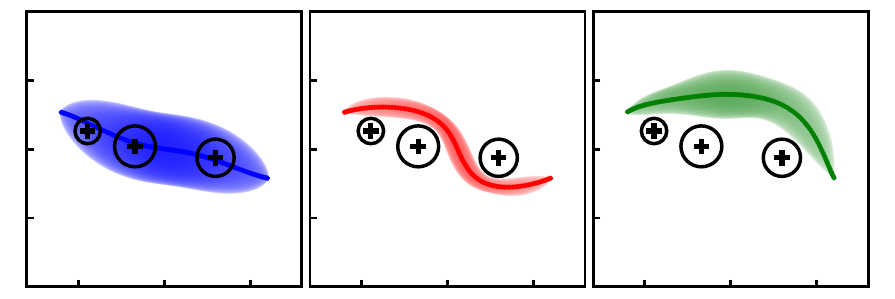}
    \caption{
    We show two examples from the randomly generated two-dimensional obstacle avoidance problems.
    The primitives are shown with a thick line, representing the mean, and a shaded area, corresponding to three standard deviations.
    The original primitive is shown in blue in the leftmost Figure, whereas the adapted primitives are shown in red (our method) and green (DEBATO) in the other two Figures.
    Obstacles are visualised as black circles with a cross at the center.
    Both methods adapt the primitive such that the the obstacles are avoided, however, our method manages to do so while staying closer to the original primitive.
    }
    \label{fig:quant_2d_marg}
\end{figure}

In this section we compare our method to DEBATO~\cite{koertDemonstrationBasedTrajectory2016}, which is a state-of-the-art approach to ProMP adaptation.
DEBATO also formalises adaptation as an optimisation problem with KL objectives, however, while our approach is based on using constraints to reshape the original ProMP, DEBATO uses additional cost functions, formally similar to our smoothness regularisation.
In addition, DEBATO uses annealing steps borrowed from relative entropy policy search~\cite{petersRelativeEntropyPolicy2010}.
In contrast to most ProMP adaptation methods, including our approach, DEBATO models trajectories as a multivariate Gaussian distribution on Cartesian coordinates and fits a ProMP after adaptation is accomplished.

Compared to other adaptation methods in the ProMP framework~\cite{colomeDemonstrationfreeContextualizedProbabilistic2017,koertLearningIntentionAware2019,osaGuidingTrajectoryOptimization2017,shyamImprovingLocalTrajectory2019}, DEBATO and our approach retain the full probabilistic characteristics when performing adaptation, meaning the end results is another ProMP.
This is a key characteristics because it allows combining and chaining different adaptations.
For these reasons we compare our method to DEBATO on two dimensional random obstacle avoidance tasks.

We define our original ProMP by sampling a start- $(-3, y_\text{start})$ and an endpoint $(+3,y_\text{end}$) and conditioning a zero mean and unit diagonal covariance on these as via-points.
We then sample one to three obstacles of varying size around the mean trajectory.
Once the initial ProMP and the obstacle avoidance task is sampled, we solve the corresponding optimisation problems without additional hyper-parameter tuning.
We choose the DEBATO parameters $B=0.001$ (KL weighting) and $\epsilon=0.001$ (REPS annealing).
We observed that these values generally did well in the random avoidance tasks.

In general the objective of an adaptation task is as following:
Find a new primitive which respects all the given constraints, while staying as close as possible to the original primitive.
For measuring this objective we use two metrics, \emph{obstacle violations} and \emph{KL}, measuring the two aspects of the task.
We define the \emph{obstacle violations} by sampling 10,000 trajectories from the adapted primitive and computing the percentage of trajectories which violate at least one of the constraints.
Therefore, the \emph{obstacle violations} metric measures whether and how well the constraints are fulfilled after adaptation.
The second metric, \emph{KL}, is computed as the KL-divergence from the adapted to the original ProMP in the weight space of the primitive.
We normalise the KL with the number of ProMP basis functions, to make this metric comparable in between experiments.
While in this specific experiment the original primitive is determined by the randomly sampled start- and endpoints, in general, we would like the adapted primitive to stay as close as possible to the original one because we assume the original primitive to contain important information about the task which we would like to preserve during adaptation.

In this experiment we first compute the raw \emph{obstacle violations} for both methods on all environments which are randomly generated as described above.
We use the same environments for both methods.
We label environments for which the raw \emph{obstacle violations} of a method exceeds 30\% as \emph{failed} for that method.
Table~\ref{tab:exp_comp_debato} shows the results of the experiment sorted by the number of obstacles in the environment.
The statistics displayed in the \emph{obstacle violations} and \emph{KL} columns are computed over all the environments for which none of the two methods failed.

The \emph{failed} columns show that our method is able to solve the two-dimensional obstacle avoidance problems reliable, failing only one percent even for the challenging three obstacle scenario, whereas DEBATO fails between $7.9\%$ and $22.0\%$ of the problems depending on the number of obstacles.
DEBATO relies on importance sampling and moment matching for adapting the primitive, which we noticed is prone to collapsing for challenging environments.
This problem might become more apparent when moving to more realistic, higher dimensional use cases like modelling the joint space primitive of a seven DOF robot arm.

For both methods we see that solved environments generally result in less than 0.5\% of sampled trajectories violating the constraints, with DEBATO averaging less violations, whereas our method is more consistent, which is reflected in a lower standard deviation.
In practice we can eliminate the possibility to sample violating trajectory by slightly enlarging the obstacle margins.
Note also that the mean trajectory is always violation free. 

Comparing the \emph{KL} metric, we observe that our method consistently manages to find solutions closer to the original ProMP.
We attribute this to the fact that we learn the Lagrange variables, such that the constraints are exactly fulfilled, during our constrained optimisation.
For DEBATO, the trade-off between KL minimization and obstacle avoidance has to be balanced by weighting these terms in the cost function, which would require a different choice of hyper-parameters for every problem.
In Figure~\ref{fig:quant_2d_marg}, we visualise two examples of generated obstacle avoidance problems that show these differences.
We can see that both methods successfully adapt the primitives such that the obstacles are avoided, however, our method manages to find solutions which stay closer to the original primitives.

\subsubsection*{Temporally unbound waypoints}
\begin{table}
  \centering%
  \caption{Evaluating temporally unbound waypoint constraints on randomly generated 2D problems}
    \begin{tabular}{lcccc}
    \toprule
    \# of waypoints & n & failed & Waypoint violations & KL \\
    \midrule
    1 & 100 & 0 (0.0\%) & 0.00\% $\pm$ 0.01\% & 0.39 $\pm$ 0.09 \\
    2 & 100 & 0 (0.0\%) & 0.01\% $\pm$ 0.03\% & 0.79 $\pm$ 0.12 \\
    3 & 100 & 0 (0.0\%) & 0.03\% $\pm$ 0.12\% & 1.18 $\pm$ 0.13 \\
    \bottomrule
    \end{tabular}
  \label{tab:exp_quant_tviaP}%
\end{table}
\begin{figure}
    \centering
    \includegraphics[width=0.49\columnwidth]{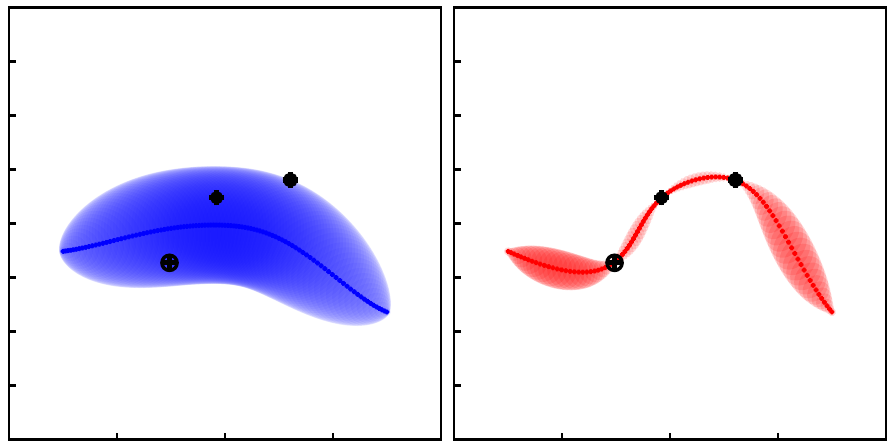}
    \includegraphics[width=0.49\columnwidth]{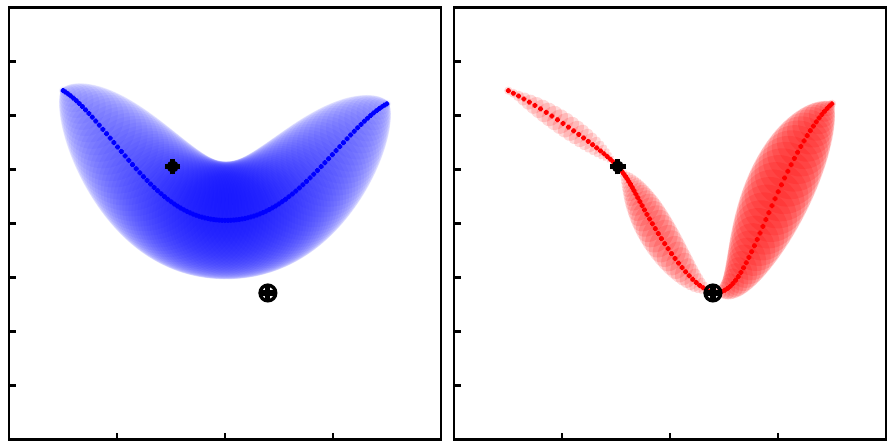}
    \caption{
      We show two examples from the randomly generated two-dimensional waypoint problems.
      The primitives are shown with a thick dotted line, representing the mean, and a shaded area, corresponding to three standard deviations.
      The original primitive is always shown in blue in the left Subfigure, whereas the adapted primitives are shown in red in the right Subfigure.
      The temporally unbound waypoints are visualised as black circles with a cross at the center.
      We can see that our method successfully adapts the primitive such that the desired waypoints are guaranteed.
    }
    \label{fig:quant_2d_marg_viaP}
\end{figure}

In this section we analyse the temporally unbound waypoint constraints on randomly generated two-dimensional environments.
We sample an initial primitive like described in the previous section~\ref{sec:exp_comp_debato}.
Then, between one and three waypoints are sampled around the mean trajectory.
Here we also ensure that the waypoints are not directly next to each other as that would result in an impossible task.
Two example problems and their solutions are depicted in Figure~\ref{fig:quant_2d_marg_viaP}.
We solve all problems with the same set of hyperparameters and show the quantitative results in Table~\ref{tab:exp_quant_tviaP}, where we use the same metrics as described in the previous Section.
We can see that our method is able to solve every single adaptation problem.
Additionally, we can verify that trajectories sampled from the adapted primitive are almost entirely violation free.
Compared to the repeller experiments in Section~\ref{sec:exp_comp_debato} we note that the normalised KL-divergence is higher, so the change to the original primitive is more significant.
This is to be expected because the waypoints force the adapted primitive to focus the probability mass on a narrow set of trajectories, whereas repellers only exclude a certain set of trajectories.

\subsubsection*{Hyperplane constraints}
\begin{table}
  \centering%
  \caption{Evaluating virtual wall constraints on randomly generated 2D problems}
    \begin{tabular}{lcccc}
    \toprule
    \# of virt. walls & n & failed & Virt. wall violations & KL \\
    \midrule
    1 & 100 & 0 (0.0\%) & 0.19\% $\pm$ 0.17\% & 0.21 $\pm$ 0.37 \\
    2 & 100 & 0 (0.0\%) & 0.27\% $\pm$ 0.19\% & 0.28 $\pm$ 0.42 \\
    3 & 100 & 0 (0.0\%) & 0.37\% $\pm$ 0.17\% & 0.37 $\pm$ 0.38 \\
    \bottomrule
    \end{tabular}
  \label{tab:exp_quant_vWall}%
\end{table}
\begin{figure}
    \centering
    \includegraphics[width=0.49\columnwidth]{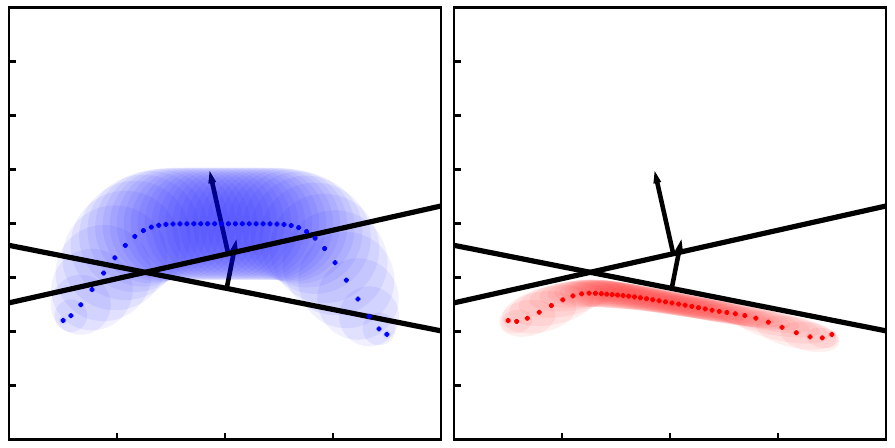}
    \includegraphics[width=0.49\columnwidth]{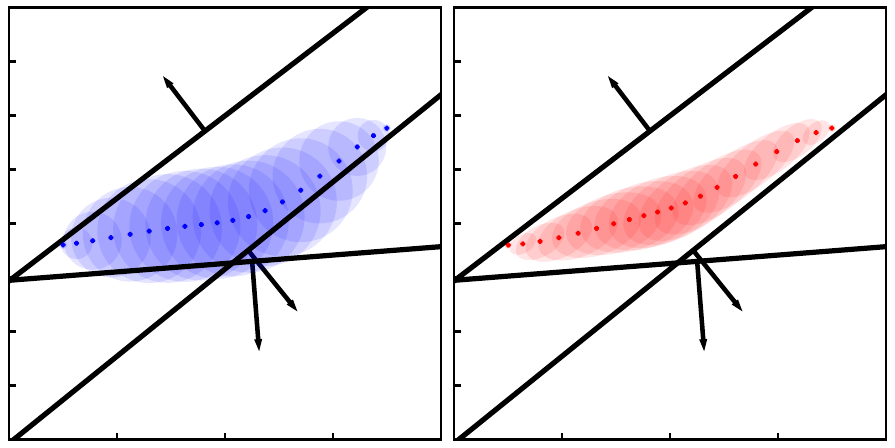}
    \caption{
      We show two examples from the randomly generated two-dimensional virtual wall problems.
      The primitives are shown with a dotted line, representing the mean, and a shaded area, corresponding to three standard deviations.
      The original primitive is always shown in blue in the left Subfigure, whereas the adapted primitives are shown in red in the right Subfigure.
      The virtual walls are visualised as black lines with the normal vector pointing in the direction in which we do not want the primitive to go.
      We can see that our method successfully adapts the primitive such that the virtual walls are avoided.
    }
    \label{fig:quant_2d_marg_vWall}
\end{figure}
  
Analogous to the previous two Sections, we investigate hyperplane constraints on two-dimensional problems.
We use the same sampling procedure to sample an initial primitive.
Then we sample between one and three hyperplanes such that the start- and end-point are allowed by the constraints.
In general, this is not a necessary condition for our method to work but it simplifies the problem.
Two example problems are shown in Figure~\ref{fig:quant_2d_marg_vWall}.
As described in the previous two Sections, all problems are solved with the same set of hyperparameters.
We present the quantitative results in Table~\ref{tab:exp_quant_vWall}.
We note that our method is able to solve all the randomly generated problems and trajectories sampled from the adapted primitive have a very high probability to be violation free.
Comparing the normalised KL-divergence we can see that the changes during adaptation are comparable to the repeller experiments and lower than during adaptation with waypoints. 

\subsection{Mutual avoidance in a dual-arm setting}
\label{SecExpPandaDualSimple}

\begin{figure}
  \centering
  \includegraphics[width=0.4\textwidth]{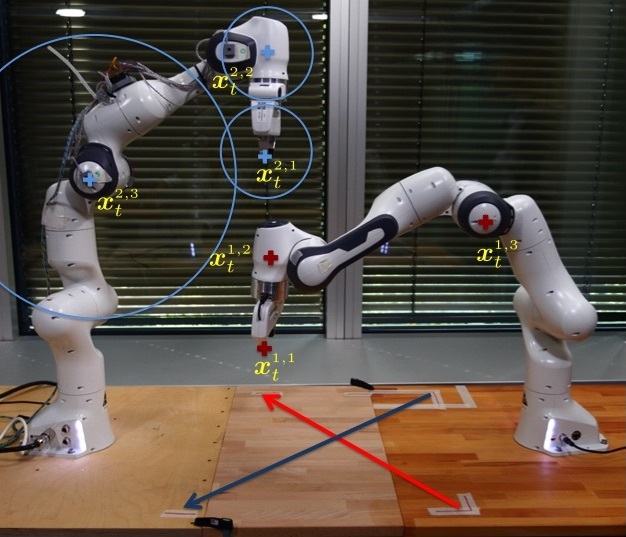}
  \caption{This Figure shows the setting of the experiments described in Section~\ref{SecExpPandaDualSimple}. The arrows display the directions of motion for the left (blue) and right (red) robot, respectively.
  The red and blue crosses display the points of interest used for the mutual avoidance constraints and the circles show the corresponding margins. 
  }
  \label{fig:dualRCross_poi}
\end{figure}
In this set of Experiments we demonstrate a range of constraints from our adaptation framework in a dual arm setting.
First, we show that the mutual avoidance constraint can be used to combine two robots with individually learned ProMPs in the same workspace.
Second, we use the joint limit constraints to fix a known problem of ProMPs.
Third, we investigate the effect of the smoothness regularisation in a real robot setting.
Finally, we examine the cross correlations of the combined primitive and we show that the adaptation encodes task-specific knowledge into the primitive.
In all experiments we use two Franka Emika Panda robots with their bases $1.13m$ apart, facing each other, as shown in Figure~\ref{fig:dualRCross_poi}.
The original task for both robots is a reaching motion, in which the end-effector has to move from the robot's left to it's right as indicated by the arrows in Figure~\ref{fig:dualRCross_poi}.
We independently demonstrate and learn separate ProMPs for each robot arm and then use our method to obtain a combined primitive with simultaneously executed, collision-free trajectories.

\subsubsection{Combining primitives in the same workspace}
For reasons of clarity we explicitly state the adaptation problem in mathematical form.
Let $p_0(\bm{w}_1)$ and $p_0(\bm{w}_2)$ be the two independently learned ProMPs and let $p(\bm{w})$ with $\bm{w} = (\bm{w}_1, \bm{w}_2)$ denote the (jointly) adapted ProMP.
In order to define constraints for collision avoidance, we choose three points of interest on each robot $\bm{x}^{\scriptscriptstyle 1,1:3}_t(\bm{w}_1)$, and $\bm{x}^{\scriptscriptstyle 2,1:3}_t(\bm{w}_2)$, respectively.
The placement of the points of interest and the distances corresponding to the mutual avoidance constraints are shown in Figure~\ref{fig:dualRCross_poi}.
One could add avoidance constraints for all possible collisions for the chosen points of interest, however, the number of constraints does not necessarily have to grow quadratically with the number of points of interest because in many practical applications a large proportion of collisions are physically improbable.
Therefore, we choose 5 of the possible 9 constraints, these are a specified by the index pairs and their corresponding distance $(i,j, d_{ij}) \in \mathcal{G}=\{(1,2,0.1),(2,1, 0.1),(2,2,0.1), (2,3,0.3), (3,2, 0.3)\}$.
Additionally, we use a hyperplane constraint with parameters $\bm{n}=(0,0,-1), b=(0,0,0.01)$ to keep the end-effectors above the surface of the table.

Given these assumptions, we formulate the resulting adaptation problem~as 
\begin{align}
  \min_{p} \: & \: D_{\textrm{\sc KL}}[\,p(\bm{w}) \vert\!\vert\, p_0(\bm{w}_1)\:p_0(\bm{w}_2)] %
  \label{EqnDualArm-1}
    \\
    \text{s.t.} \: & \:  P_{\bm{w}}( \vert \bm{x}^{1,i}_{t}(\bm{w}_1) -  \bm{x}^{2,j}_{t}(\bm{w}_2) \vert^2 > d_{ij}^2  ) \geq \alpha_{\mathrm{c}} , 
    \:  (i,j, d_{ij}) \in \mathcal{G}
  \nonumber
  \\
  \: & \:   P_{\bm{w}}\Bigl(\bm{n}^{T}\bigl(\bm{x}^{i,1}_t(\bm{w}_i)-\bm{b}\bigr) \leq 0\Bigr) \geq \alpha_{\mathrm{hp}}, \quad  i=1,2
  \nonumber
\end{align}
where we set $\alpha_{\mathrm{c}} = \alpha_{\mathrm{hp}} = 0.999$.
Optimisation is carried out with Algorithm~\ref{alg:cpmp}.

\label{sec:exp_dualCross_basis}
\begin{figure}
  \centering
  \includegraphics[width=0.48\textwidth]{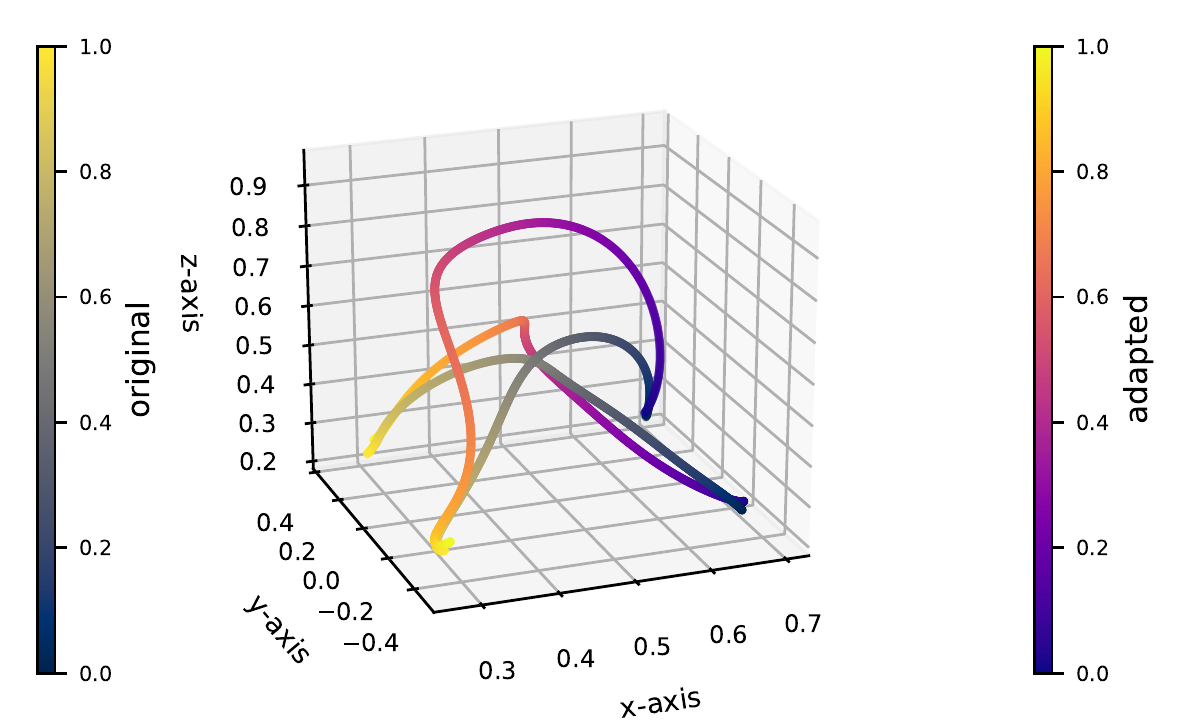}
  \caption{In this Figure we show the mean trajectories of the points of interest $x_t^{1,2}$ and $x_t^{2,2}$ (see Figure~\ref{fig:dualRCross_poi}) before (original) and after adaptation.
  Time is indicated by the colour gradients. If executed the original primitives would lead to a collision of the robots, whereas the adaptation shapes the trajectories to be collision-free.}
  \label{fig:dualRCross_3d}
\end{figure}
The mean trajectories of the original ProMP are plotted in Figure~\ref{fig:dualRCross_3d} showing that the end-effectors would collide if the trajectories were executed simultaneously.
The adaptation has a stronger effect on the motion of the left robot resulting in trajectories that move the end-effector on top of the right robot.
Additionally, the adaptation shapes both trajectories in a way that increases the distance between both robots. 
The video in the Supplementary Material shows several samples from the adapted primitive.

This experiment also shows the distinct advantage of probabilistic approaches for primitives over their deterministic counterparts.
The probabilistic approach allows the adaptation to have a stronger effect on parts of the trajectory that have a larger variance and thus are  less essential for the reaching task. 
In the following we introduce various modifications to the this mutual avoidance experiment in order to examine specific details of our method.
\subsubsection{Respecting joint limits}
\label{sec:exp_dualCross_jLimits}
\begin{figure}
  \centering
  \includegraphics[width=0.45\textwidth]{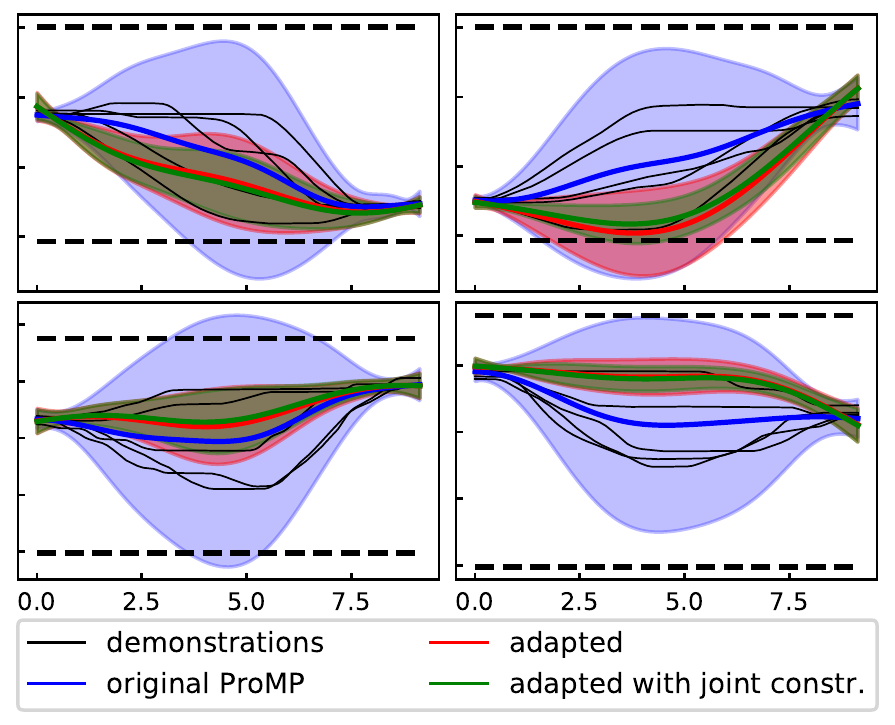}
  \caption{We visualise the trajectories of joints 4 and 6 in experiment~\ref{sec:exp_dualCross_jLimits}. We show the demonstrations (black) and the mean with $\pm 3$ standard deviations of the primitive before (blue) and after adaptation (red and green).
  Solving the adaptation problem described in eq.~\eqref{EqnDualArm-1} results in the primitive shown in red, whereas the primitive shown in green uses additional limit constraints at the joint limits (dashed black).
  Even though all demonstrations stay within the joint limits, both the original ProMP and the ProMP adapted without limit constraints can have mass outside of the feasible limit.}
  \label{fig:dualRCross_jLimits}
\end{figure}

\begin{figure*}
  \subfloat[Smoothness Regularisation: Joint Space marginal distribution]{
    \includegraphics[width=0.53\textwidth]{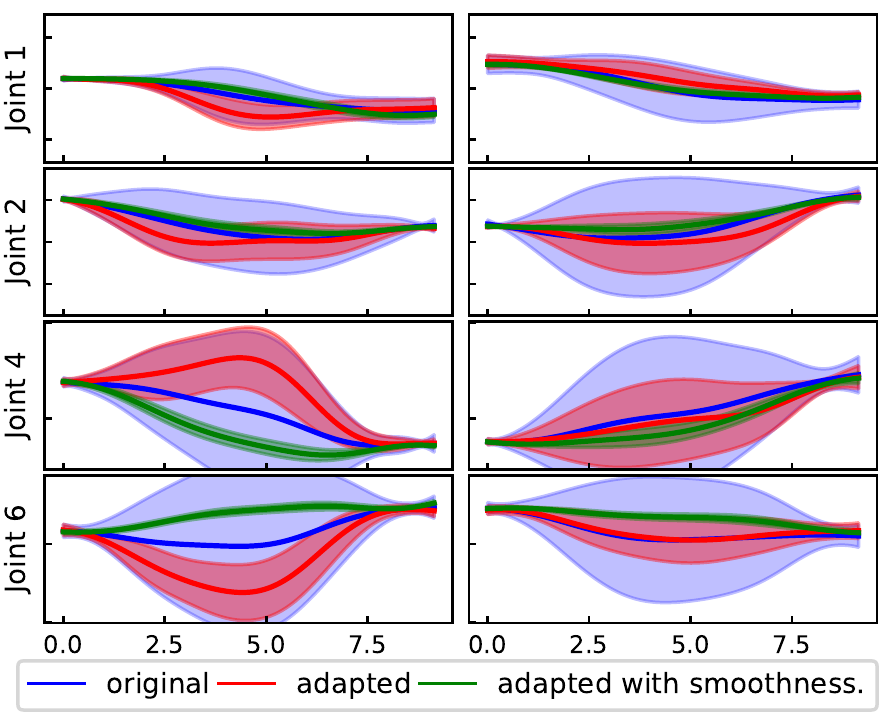}
    \label{sfig:dualRCross_smooth_jMarginals}
  }%
  \hfil%
  \subfloat[Smoothness Regularisation: Snapshots]{
    \includegraphics[width=0.38\textwidth]{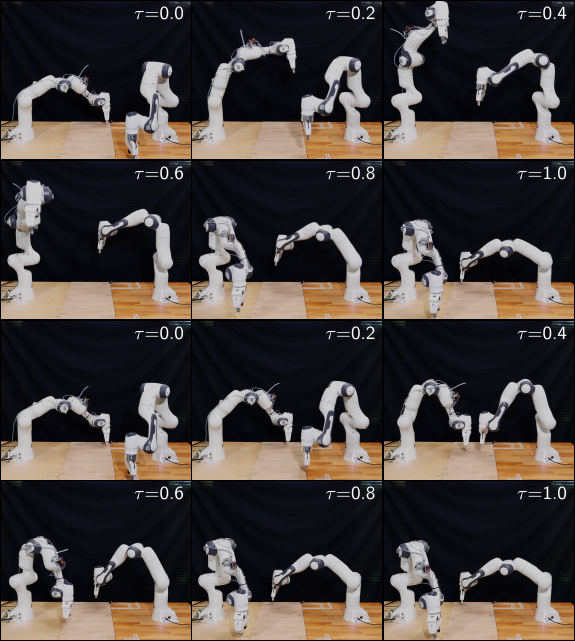}
    \put(-209,135){\rotatebox{90}{\footnotesize without smoothness}}
    \put(-209,25){\rotatebox{90}{\footnotesize with smoothness}}
    \label{sfig:dualRCross_snapshots}
  }%
  \caption{This Figure demonstrates the effect of smoothness regularisation.
  On the left we visualise the joint space marginals with mean and $\pm 3$ standard deviations.
  We compare the original ProMP (blue) to the ones adapted with (green) and without (red) smoothness regularisation.
  The right Figure shows snapshots of the mean trajectories, with (bottom two rows) and without (top two rows) smoothness, executed on the robots.
  We can see that the smoothness regularisation significantly reduces the remaining variance and results in a smoother trajectory.
  Instead of moving the left robot on top of the right one, the smooth trajectories use the available space between the two robots by moving both of them closer to their own bases. 
  }
  \label{fig:dualRCross_smoothness}
\end{figure*}
A side-effect of modelling with ProMPs is that even though all demonstrated movements stay within the robot's joint limits, the ProMP can have mass outside of the limits.
Accordingly, we can apply the joint space constraints to prevent joints going off-limits in the adapted ProMP. 
In Figure~\ref{fig:dualRCross_jLimits}, we visualise the application of limiting the ProMP in joint space directly, as described in Section~\ref{sec:method_jconst_limits}. 
Observe how adaptation with additional joint limits, shown on green, removes joint limit violations when compared to the original (blue) and adapted (red) ProMPs.

\subsubsection{The effect of Smoothness Regularisation}
\label{sec:exp_dualCross_smooth}

We compare adaptation with and without smoothness regularisation by adding a regularisation term $ \kappa \,E_{p(\bm{w})}[R(\bm{w})]$ with $\kappa=0.1$ to \eqref{EqnDualArm-1} without any additional joint weighting.
Figure~\ref{sfig:dualRCross_smooth_jMarginals} shows the marginal trajectories for a few relevant joints while Figure~\ref{sfig:dualRCross_snapshots} compares snapshots of the resulting mean trajectories.
We can observe that smoothness regularisation reduces the variance and results in a significantly different trajectory distribution.
Similarly to the toy example in Figure~\ref{fig:joint_smooth}, the adapted trajectory exhausts the margins given by the constraints to maximise smoothness.
This is particularly evident when comparing the trajectories visualised in Figure~\ref{sfig:dualRCross_snapshots}: instead of moving one robot on top of the other, the mean trajectory shown in the bottom two rows barely keeps the enforced distance but is much more economical.
The video in the Supplementary Material shows the different trajectories executed side-by-side for a better comparison. 

\subsubsection{Adaptation encodes task specific knowledge}
\label{sec:exp_dualCross_correl}
\begin{figure*}
  \subfloat[Via-point conditioning: 3D overview]{
    \includegraphics[width=0.4\textwidth]{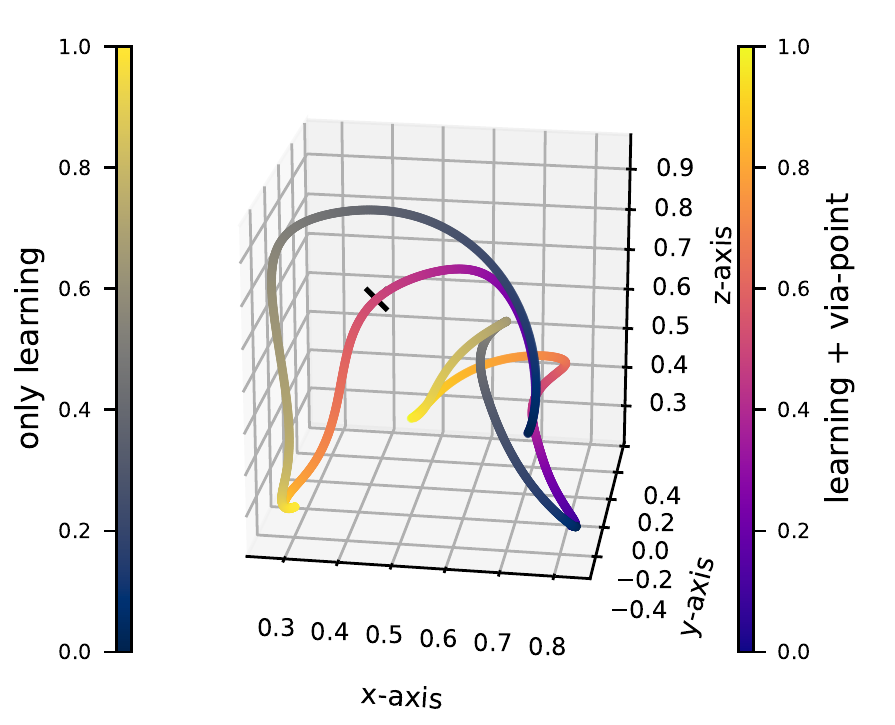}
    \label{sfig:dualRCross_viaPoint_3d}
  }%
  \hfil%
  \subfloat[Via-point conditioning: Cartesian marginal distribution]{
    \includegraphics[width=0.4\textwidth]{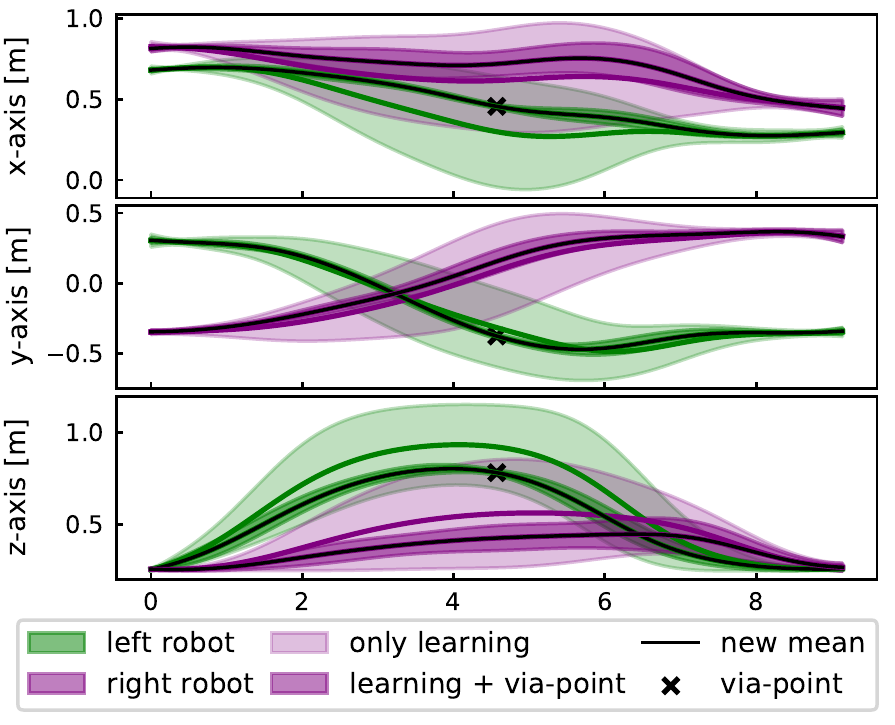}
    \label{sfig:dualRCross_viaPoint_cMarginals}
  }
  \caption{%
    This Figure accompanies the experiment on encoding task specific knowledge during adaptation, which is described in Section~\ref{sec:exp_dualCross_correl}.
    The left Figure shows the mean Cartesian trajectories of the points $x_t^{1,2}$ and $x_t^{2,2}$ (see Figure~\ref{fig:dualRCross_poi}) on both robots once after adaptation (only learning) and once after we conditioned the adapted ProMP on a via-point (learning + via-point).
    Time is indicated by the colour gradients and the black "x" in both images represents the joint space via-point transformed through the forward kinematics.
    Figure~\ref{sfig:dualRCross_viaPoint_cMarginals} shows the corresponding Cartesian marginal distributions, also before (low alpha) and after (high alpha) conditioning.
    We can clearly see that conditioning the left robot also significant changes the trajectory of the right robot, that is, the right robot moves further away from the path of the left robot.
  }%
  \label{FigDualArmSimpleViaSmooth}
\end{figure*}
An important advantage of the probabilistic approach is that it can capture covariances of the robot's motion.
These covariances exist not only between individual joints of the same robot, but they also link the joints of different robots.
This implies that task specific knowledge, like the robots avoiding each other, can be encoded in the covariance matrix $\bm{\Sigma}_w$.
To test this hypothesis we designed the following experiment: we condition the adapted ProMP, using the KL-only objective from Section~\ref{sec:exp_dualCross_basis}, on a via-point in the joint space of the left robot to measure the effect it has on the right robot's trajectories.
We place the via-point in the joint space and force the left robot to move through the Cartesian position shown in Figure~\ref{sfig:dualRCross_viaPoint_3d}. At the same time we observe how this adaptation changes the trajectory of the right robot.
In this experiment we use the sum of marginal KL's objective $D_{\textrm{\sc KL}}[\,p(\bm{w}_1) \vert\!\vert\, p_0(\bm{w}_1)]  + D_{\textrm{\sc KL}}[\,p(\bm{w}_2) \vert\!\vert\, p_0(\bm{w}_2)]$ (see Sections~\ref{sec:method_task_dualAvoidance} and~\ref{SecExpPlanarDual}), to incentivise stronger correlations between the two robots.
The via-point conditioning results in the left robot's trajectory moving closer to the right robot, both in x and z direction, as is shown in Figure~\ref{sfig:dualRCross_viaPoint_cMarginals}.
Furthermore, the figure shows the right robot adapting its trajectory accordingly, by moving closer to the table and further away from the left robot's base.
We can observe that, as expected, the right robot's trajectory has changed in a suitable way to reduce the likelihood of a collision, thus correlations in the adapted ProMP do indeed encode avoidance to a measurable degree.
However, note that by conditioning we do not obtain guaranteed collision avoidance since avoidance is not enforced by a constraint.
We also want to emphasise that this task specific knowledge must have been added during the adaptation with our framework, as the original ProMP was learned from individual motions.
The video in the Supplementary Material also visualises this experiment and shows an image of the ProMP correlations before and after adaptation.

In this set of experiments we showed how joint adaptation of ProMPs can be carried out in a real world scenario.
This type of adaptation can be useful in many practical applications where several robots, each having their own independently trained tasks, can be adapted and reconfigured to operate in the same environment.
We also showed that the adaptation process can add task specific knowledge to the ProMP, making downstream adaptations, like conditioning, have a more informed effect.

\subsection{Mutual avoidance with unbound waypoints}
\label{SecExpPandaDualHard}
In the previous experiment we showed how to combine multiple individual primitives into a combined motion, allowing us to reconfigure robots in close proximity.
However, the adaptation could find collision-free trajectories by just coordinating the robots spatially, because the key points both robots had to visit---the start and end location of their motions---were naturally distinct for both robots.
In manufacuring tasks this setup can be violated because multiple robots might have to visit the same location requiring not only spatial, but also temporal coordination.
Inspired by a pick-and place task, in which two robots have to grasp objects out of the same box, we designed an experiment where we require the two robots end-effectors $\bm{x}^{1}_{t}(\bm{w}_i)$ and $\bm{x}^{2}_{t}(\bm{w}_i)$ to visit the same location in Cartesian space without colliding.
Compared to the previous adaptation, this is a somewhat more challenging problem requiring us to specify the task relevant parameters with constraints of the form: visit a specific location at least once during the trajectory while staying away from the other robot all the time.
This task requires us to combine collision avoidance with temporally unbound waypoints, described in Section~\ref{sec:method_task}.
To demonstrate this adaptation task we used the dual arm setup shown in Figure~\ref{sfig:dualRBox_photo}.
The robots are placed next to each other with parallel $x$-axes and are required to perform a round-trip trajectory to the middle of the table.
Similarly to the experiment presented in the previous section, we demonstrate and learn two independent ProMPs $p_{0}(\bm{w}_{1})$ and $p_{0}(\bm{w}_{2})$ and then adapt them jointly.

To formulate the adaptation problem, we use the following constraints for the end-effectors $\bm{x}^{1}_{t}(\bm{w}_1)$  and $\bm{x}^{2}_{t}(\bm{w}_2)$: (i)~a mutual avoidance constraint; (ii)~temporally unbound waypoints for both end-effectors at location $ \bar{\bm{x}}_{\mathrm{middle}}$; (iii)~waypoint constraints for the start and end locations $\bar{\bm{x}}^{i}_{0}$ and $\bar{\bm{x}}^{i}_{T}$; (iv)~a hyperplane constraint to keep both end-effectors above the surface of the table.
By using a similar notation like in the previous section, the adaptation problem is formalised as
\begin{align}
 	\min_{p} \: & \: D_{\textrm{\sc KL}}[\,p(\bm{w}) \vert\!\vert\, p_0(\bm{w}_1)\:p_0(\bm{w}_2)] + \kappa \,E_{p(\bm{w})}[R(\bm{w})]
	\label{EqnDualArm-2}
  	\\
  	\text{s.t.} \: & \:  P_{\bm{w}}\bigl( \vert \bm{x}^{1}_{t}(\bm{w}_1) -  \bm{x}^{2}_{t}(\bm{w}_2) \vert^2 > d_{\mathrm{coll}}^2  \bigr) \geq \alpha_{\mathrm{coll}} 
	\nonumber
	\\
  	\: & \:  \max\limits_{t} P_{\bm{w}}\bigl( \vert \bm{x}^{i}_{t}(\bm{w}_i) -  \bar{\bm{x}}_{\mathrm{middle}} \vert^2 \leq d^2_{\mathrm{middle}}  \bigr) \geq \alpha_{\mathrm{middle}} 
	\nonumber
	\\
  	\: & \: P_{\bm{w}}\bigl( \vert \bm{x}^{i}_{t=0}(\bm{w}_i) -  \bar{\bm{x}}^{i}_{0} \vert^2 \leq d^2_{i,0}  \bigr) \geq \alpha_{0} 
	\nonumber
	\\
  	\: & \:  P_{\bm{w}}\bigl( \vert \bm{x}^{i}_{t=T}(\bm{w}_i) -  \bar{\bm{x}}^{i}_{T} \vert^2 \leq d^2_{i,T}  \bigr) \geq \alpha_{T} 
	\nonumber
	\\
	\: & \:   P_{\bm{w}}\Bigl(\bm{n}^{T}\bigl(\bm{x}^{i}_t(\bm{w}_i)-\bm{b}\bigr) \leq 0\Bigr) \geq \alpha_{\mathrm{hp}}, \quad  i=1,2.
	\nonumber
\end{align}
We set $\alpha_{\mathrm{coll}} , \alpha_{\mathrm{middle}}, \alpha_{0}, \alpha_{T} $ and $\alpha_{\mathrm{hp}}$ to $0.999$ and use Algorithm~\ref{alg:cpmp} to solve the optimisation problem.
The resulting adapted ProMP is illustrated in Figure~\ref{FigDualArmBox} and in the supplementary Video.
Since both end-effectors have to visit $\bar{\bm{x}}_{\mathrm{middle}}$, collision avoidance happens through the left arm delaying and the right arm speeding up its motion towards $\bar{\bm{x}}_{\mathrm{middle}}$. 
As we can observe in Figure~\ref{sfig:dualRBox_joints}, the left arm moves slowly during the first part of the trajectory followed by fast movement during the second part.
The right arm displays a complementary motion pattern.
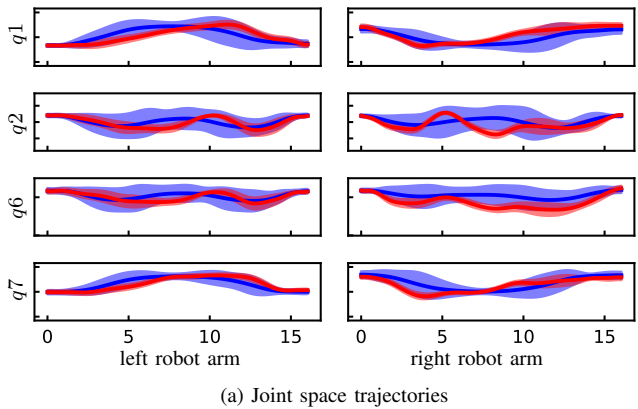
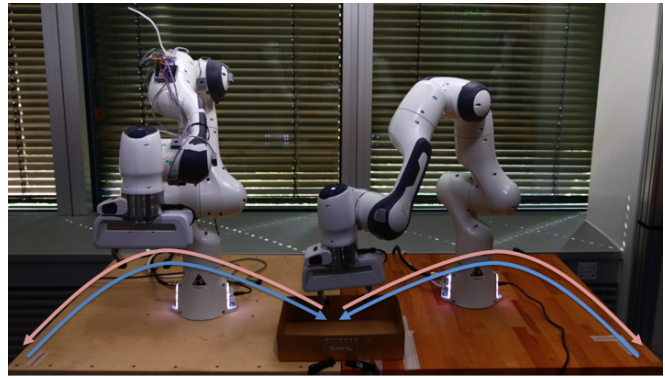
\begin{figure*}[t]
    \centering
    \subfloat[Joint space trajectories]{
        \includegraphics[width=0.48\textwidth]{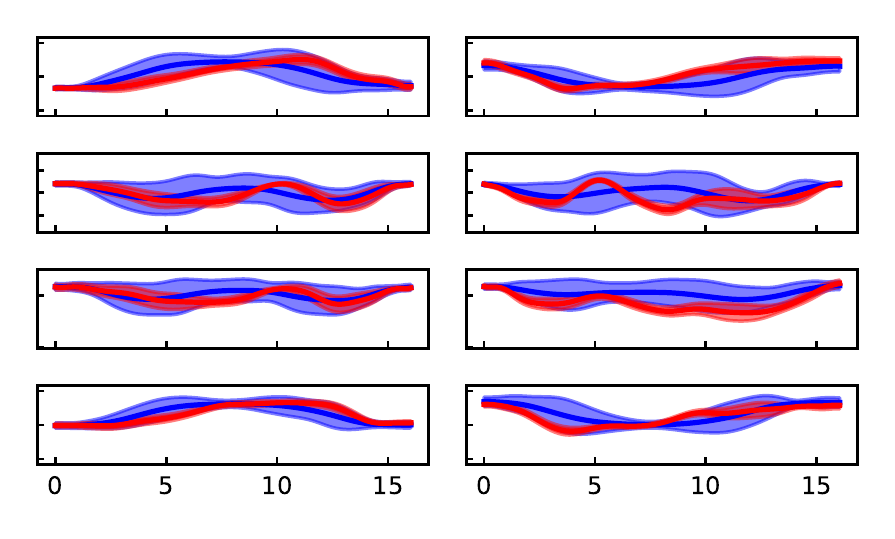}
        \put(-205,4){\footnotesize left robot arm}
        \put(-95,4){\footnotesize right robot arm}
        \put(-247,125){\rotatebox{90}{\footnotesize$q1$}}
        \put(-247,91){\rotatebox{90}{\footnotesize$q2$}}
        \put(-247,59){\rotatebox{90}{\footnotesize$q6$}}
        \put(-247,28){\rotatebox{90}{\footnotesize$q7$}}
        \label{sfig:dualRBox_joints}
    }
    \hfil
    \subfloat[Experimental setup.]{
        \includegraphics[width=0.48\textwidth]{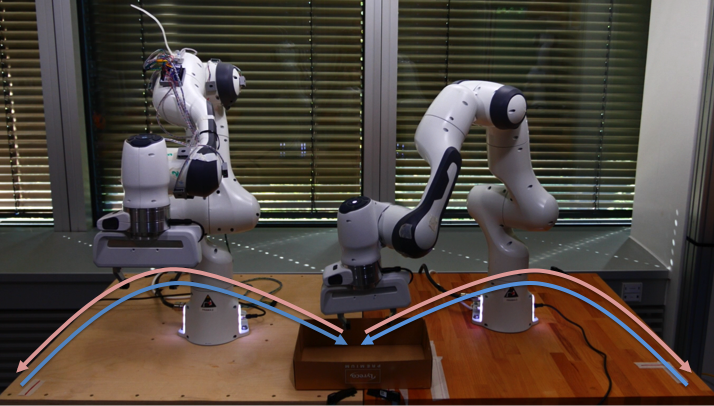}
        \label{sfig:dualRBox_photo}
    }
    \caption{\emph{Illustrating a mutual avoidance task with unbound waypoints.} The panel (b) illustrates the experimental setup for the mutual avoidance task presented in Section~\ref{SecExpPandaDualHard}. The arrows show the first (blue) and second (red) parts of the original ProMP trajectories together with a snapshot of the adapted trajectories (background).
    Figure (a) shows the relevant joint space trajectories, joints 1, 2, 6 and 7, of both robots.
    The original and adapted ProMPs are shown in blue and red, respectively.
    Adaptation results in the left arm delaying (see flat sections in the first half of the trajectory) and the right arm speeding up its movement towards the middle.}
    \label{FigDualArmBox}
\end{figure*}

While in the previous example the constraint allowed a large variety of trajectories without collision,  in this example, avoiding collision requires a much finer temporal coordination.
We believe that this is due to the strict geometric constraints and the temporal dependencies prompted by the unbound waypoint.

In this experiment we demonstrated that highly non-trivial path constraints such as the unbound waypoint can be implemented by the proposed adaptation method.
This constraint can be further generalised by requiring the arm to spend a certain amount of time at the unbound waypoint.
We can then use this time to perform some additional task which is not necessarily part of the ProMP, say, a pick or drop task.

\clearpage{}%

\clearpage{}%
\section{Related Work}
\label{sec:related}

Adapting movement primitives to new scenarios is a central element in every primitive framework.
In this section we discuss the related work on primitive adaptation and we draw connections to our own approach.

\subsection{Adaptation in other primitive frameworks}
DMPs formulate primitives in terms of dynamic systems, which allows a convenient extension to obstacle avoidance by adding repellent points to the spring-damper behaviour~\cite{parkMovementReproductionObstacle2008}.
This concept has been extended to multiple obstacles~\cite{hoffmannBiologicallyinspiredDynamicalSystems2009}, \hl{volumetric obstacles~\cite{ginesiDynamicMovementPrimitives2019,chiLearningGeneralizationObstacle2019}} and the authors of~\cite{krugModelPredictiveMotion2015} proposed an online model predictive controller based on DMPs for tackling online obstacle avoidance.
In our approach we also presented repellent points as one way to achieve obstacle avoidance.
Specifically, the direct probabilistic encoding of trajectories allows us to formulate repellers in terms of excluding a specific region in space from the primitive.
\hl{This has the advantage that we can explicitly specify a margin instead of tuning the repellent force, an advantage we share with work on volumetric obstacles for DMPs~\cite{ginesiDynamicMovementPrimitives2019}.}
Furthermore, we can exploit the unscented transform to add repellent points to any point of interest on the robot.

Another extension of the DMP framework incorporates joint limits~\cite{duanConstrainedDMPsFeasible2018}.
This extension is based on transforming the DMP into a space in which the joint limits correspond to $\pm \infty$~\cite{charbonneauOnlineJointLimit2016}.
In Section~\ref{sec:method_jconst_limits} we discussed how ProMPs could use the same technique for joint limit avoidance.
Besides joint limits, our method can also operate with time varying limits and demonstrations that do not fulfil the limits in the first place. 
In~\cite{dahlinAdaptiveTrajectoryGeneration2020}, the authors modified the dynamic system governing the DMP's evolution of time to guarantee staying within velocity limits without perturbing the trajectory.
In our approach joint velocity limits could be handled with the limit constraints presented in Section~\ref{sec:method_jconst_limits}, however one has to encode the joint velocity directly in the ProMP.
Moreover, limits on Cartesian velocity could be handled analogously to hyperplane constraints, relying on the robot's Jacobian to transform the ProMP linearly into Cartesian velocity space.
Finally, the combination of multiple DMPs has been studied in~\cite{loberMultipleTaskOptimization2014}.
Similar to our approach, the authors use constrained optimization, in their case to incorporate the robot's equations of motion. However, in theory their formulation allows constraints on joint torques, joint acceleration and wrench forces.
\hl{
  Recent work also proposed probabilistic formulations of the DMP framework, by allowing the forcing function to be represented by a distribution~\cite{meierProbabilisticRepresentationDynamic2016} and DMPs have been extended to deal with linear constraints in~\cite{saverianoLearningBarrierFunctions2019}.
}

The GMM-GMR framework can formulate adaptation by combining multiple primitives, representing different skills, into one.
In~\cite{calinonStatisticalLearningImitation2009} the authors combine task space and joint space primitives.
They use linear, Jacobian-based transformations to map different primitives into the same space in which they are merged using Gaussian multiplication.
Combination through Gaussian multiplication is the basis for a range of adaptation techniques in the framework: Task parametrized Gaussian mixture models (TP-GMM)~\cite{calinonTutorialTaskparameterizedMovement2016} use different linear transformations to record the same primitive from different frames.
After learning, adaptation can be done by changing the transformation function to generalise a primitive to, for example, unseen start- or end-points.
In~\cite{silverioBimanualSkillLearning2018} TP-GMM is extended to include frames representing Cartesian orientation and the authors apply their method to a bimanual setting.
The same authors develop an approach for combining primitives based on learned task hierarchies in~\cite{silverioLearningTaskPriorities2019}.
Similar to our method, the aforementioned approaches allow adaptation in both joint and task-space.
In Section~\ref{sec:method_combining} we sketched how a combination of primitives could be performed in our framework, however, the main focus of our paper lies on using constraints to formulate adaptation in terms of including and excluding behaviour from learned primitives.
\hl{
  In~\cite{huangGeneralizedTaskParameterizedSkill2018}, the authors use reinforcement learning to adapt the TP-GMM framework for Cartesian obstacle avoidance.
}

\hl{
  The KMP framework has been extended for adaptations by including linear constraints~\cite{huangLinearlyConstrainedNonparametric2020} as well optimising for smoothness of the trajectory~\cite{huangOrientationLearningAdaptation2020}.
  In~\cite{huangLinearlyConstrainedNonparametric2020} the authors also formulate adaptation as a constrained optimization problem, in which the constraints can for example be used to add virtual walls or project the motion onto a hyperplane.
  The formulation leads to a quadratic cost function with linear constraints, which can be quickly solved with quadratic programming (QP).
  Compared to our approach there are two major differences:
  First,~\cite{huangLinearlyConstrainedNonparametric2020} optimizes for an optimal parameter vector $\bm{w}^{*}$, whereas we consider $\bm{w}$ as a random variable, thus we optimize for both, mean and variance.
  Second, the formulation only allows linear constraints, which forces the primitive to be described in the same space as the linear constraints.
  In case of virtual walls, the primitive has to be a Cartesian space primitive, whereas joint limits require a joint space primitive, making it impossible to combine Cartesian space and joint space adaptation techniques.
  Compared to that, our approach also allows adding constraints on multiple points of interest, such as the end-effector and the elbow at the cost of a higher computational complexity.
  Increasing the smoothness of a KMP is considered in~\cite{huangOrientationLearningAdaptation2020}.
  Similar to our approach, the authors propose to add a penalty on the second-order derivative of the basis functions, however this penalty is only used to find an optimal parameter vector $\bm{w}^{*}$, whereas our approach adapts both mean and variance when optimising the smoothness of a primitive as detailed in Section~\ref{sec:method_jconst_smooth}.
}

\subsection{Adaptation in the ProMP framework}
In~\cite{paraschosProbabilisticPrioritizationMovement2017} the authors formulate the combination of multiple ProMPs based on Gaussian multiplication.
In the same work, the author tackles obstacle avoidance by adding a primitive which explicitly moves around an added obstacle.
In our framework obstacle avoidance is best handled with a repellent point, however, the combination of different ProMPs based on Gaussian multiplication could be used before or after adaptation to exploit the redundancy of a manipulator and solve multiple tasks in parallel.

A different approach to obstacle avoidance, based on trajectory optimization with repellent points, has been presented in~\cite{koertDemonstrationBasedTrajectory2016}.
The trajectory distribution is modelled as a Gaussian on a discretisation of the Cartesian path.
This distribution is optimised to stay as close as possible to a given ProMP---also in the Cartesian space---while maximising a reward function which incentivises obstacle avoidance.
After optimisation, samples from the trajectory distribution are used to learn a new (adapted) Cartesian space ProMP.
This ProMP is then subsequently used for online via-point conditioning as presented in~\cite{paraschosProbabilisticMovementPrimitives2013}.
In comparison, our approach can be used with ProMPs which directly parametrise joint space trajectories and we formulate obstacle avoidance as constraints thus removing the need for tuning a reward function.
Additionally, we can achieve obstacle avoidance for multiple robot links, by using the unscented transform combined with the robot's forward kinematics.
\hl{
  Constrained optimisation of KL-divergence objectives has also been recently used in other fields, such as machine learning~\cite{rezendeTamingVAEs2018,klushynLearningHierarchicalPriors2019}.
}

In~\cite{colomeDemonstrationfreeContextualizedProbabilistic2017,koertLearningIntentionAware2019} the authors use ProMPs as a basis for online obstacle avoidance.
They formulate a deterministic approach to obstacle avoidance by minimising the Mahalanobis distance to the original ProMP, while at the same time keeping a defined distance to an obstacle. This has the benefit of requiring fewer parameters, making the method realtime feasible.
However, having a distribution over trajectories allows us to learn useful correlations, between individual joints as well as between different robots, allowing improved adaptation afterwards, as we showed in Section~\ref{sec:exp_dualCross_correl}.  
In general, our approach can be used to pre-adapt a given ProMP to new situations, with online obstacle avoidance as described in~\cite{colomeDemonstrationfreeContextualizedProbabilistic2017,koertLearningIntentionAware2019} being used afterwards to react to dynamically changing conditions.

In~\cite{gomez-gonzalezUsingProbabilisticMovement2016} and in a follow-up work~\cite{gomez-gonzalezAdaptationRobustLearning2020} the authors show how to adaptively condition a ProMP learned in joint space on a desired end-effector position, orientation and velocity.
They present a table-tennis task where the robot's striking motion has to be dynamically conditioned on the predicted hitting point of a moving ball.
The approach relies on a first order expansion of the robot's forward kinematics and uses a Laplace approximation to find a Gaussian posterior for conditioning.
This is a particularly well suited method for fast real-time adaptation with Cartesian via-points, it can also be used in combination with a ProMP adapted by our method.

\hl{
  In a different set of works, ProMPs are used to improve planning algorithms, specifically CHOMP, by adding the mahalanobis distance to a learned primitive as an additional cost term~\cite{osaGuidingTrajectoryOptimization2017,shyamImprovingLocalTrajectory2019}. Similar to our method these approaches try to solve a problem in which they want to stay as close as possible to a given primitive, while avoiding obstacles and minimising smoothness. However, different to our approach,~\cite{osaGuidingTrajectoryOptimization2017} and~\cite{shyamImprovingLocalTrajectory2019} optimize for a single trajectory, that solves the specific task, whereas we try to find a new, adapted primitive, that is a distribution over trajectories.
  This has the advantage that we can chain different adaptations and build libraries of primitives.
}

\clearpage{}%

\clearpage{}%
\section{Conclusions}
In this paper we introduce a unified probabilistic framework for adapting Probabilistic Movement Primitives to new scenarios.
ProMPs can be conveniently learned from demonstrations and they encode relevant information about important aspects of the task in their variance. 
We formulate adaptation as a constrained optimisation problem where we constrain the probability mass associated with undesired trajectories to be low, while retaining the probabilistic representation.
Our framework accommodates a variety of constraints which can be used as tools to shape the distribution over trajectories both in joint and task space.
This enables us to formulate and solve a rich class of adaptation problems, such as, imposing limits in joint space, adding virtual walls in Cartesian coordinates, avoiding obstacles or placing several robots with individual primitives in the same workspace, by combining several adaptations in a single constrained optimisation problem.
Compared to previous cost-function based approaches our method does not require hyper-parameter tuning because the different costs (constraints) are automatically balanced while fitting the Lagrange multipliers.
  
There are several important immediate practical applications of our approach.
In Section~\ref{sec:method_jconst_smooth}, we introduce smoothness regularisation that can be used as an ad-hoc tool for regularisation when one is agnostic about the number of basis functions to use in the ProMP.
The resulting smooth trajectories are also easier to control.
In Section~\ref{SecExpPandaDualSimple}, we demonstrate how the inherent problem of ProMPs in modelling bounded joints with unbounded Gaussian variables can be addressed and partially rectified by using joint limiting constraints.
The mutual avoidance adaptation presented in Sections~\ref{SecExpPandaDualSimple} and~\ref{SecExpPandaDualHard} show how our framework can be successfully applied to reconfigure robots with independently learnt tasks to jointly operate in the same workspace. 
Additionally, we showed in Section~\ref{sec:exp_dualCross_correl} how post-optimisation adaptations can have a more informed effect, due to the addition of task specific knowledge to the primitive. 

Retaining a full probabilistic approach in adaptation comes with the disadvantage of having a large number of parameters---a quadratic number of parameters when compared to the deterministic approaches.
However, we can reduce the number of covariance parameters (e.g. low-rank and sparse structures~\ref{SecOptimisation}) and make our approach online feasible.
The focus of our current work is establishing a unifying framework for ProMP adaptation, we intend to extend our approach to online adaptation in a future work.
\clearpage{}%

\endgroup

\section*{Acknowledgment}

The authors would like to thank their colleagues at the Volkswagen Group Machine Learning Research Lab for their invaluable suggestions towards improving the manuscript and for their contributions to the robotics setup used in this work.

\ifCLASSOPTIONcaptionsoff
  \newpage
\fi

\bibliographystyle{IEEEtran}
\bibliography{ctl,lit}
\begin{IEEEbiography}[{\includegraphics[width=1in,height=1.25in,clip,keepaspectratio]{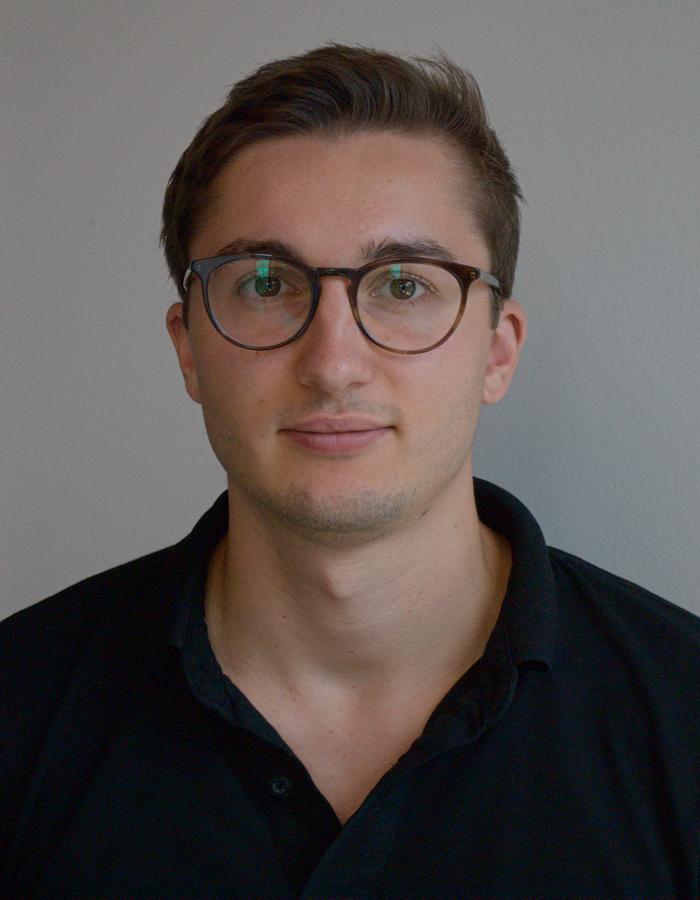}}]{Felix Frank}
received his MSc degree in automation and control from RWTH Aachen University. He has worked as a research student in the field of optimal control for vehicle combustion engines. Felix joined the Volkswagen Group Machine Learning Research Lab in 2017 and he is currently pursuing a Ph.D. degree on topics related to stochastic optimal control and reinforcement learning in robotics.
\end{IEEEbiography}
\begin{IEEEbiography}[{\includegraphics[width=1in,height=1.25in,clip,keepaspectratio]{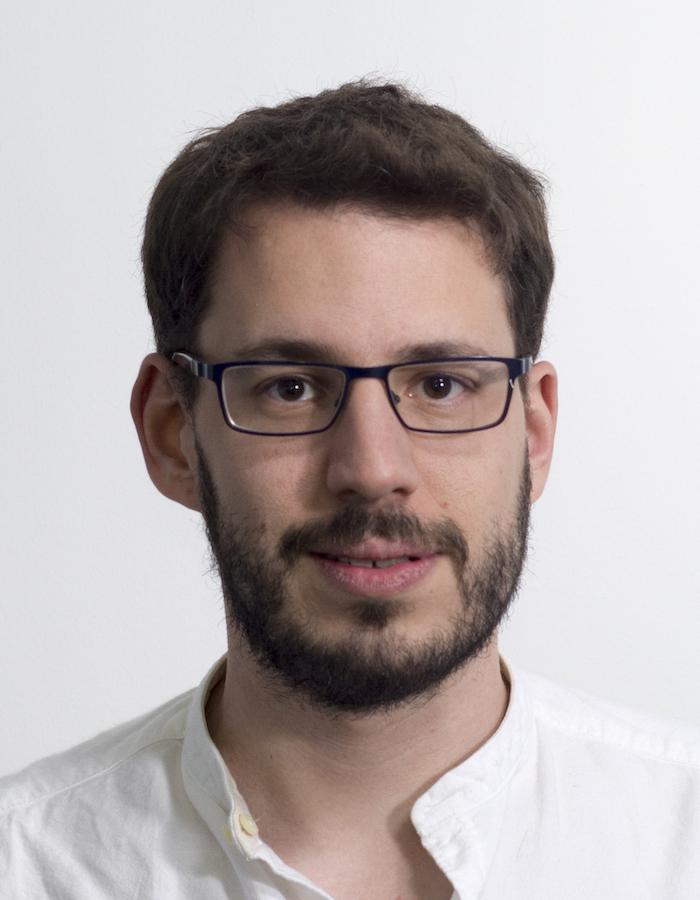}}]{Alexandros Paraschos}
received his Ph.D  degree in  computer science  from Technical University of Darmstadt and is working in the areas of robotics and machine learning. During his PhD, he focused on Robot Learning for Complex Motor Skills. Before his PhD, Alexandros has been a research associate in Cognitive Robotics Research Centre (CRRC), at University of Wales. In 2017 he joined the Volkswagen Group Machine Learning Research Lab as a research scientist.
\end{IEEEbiography}
\begin{IEEEbiography}[{\includegraphics[width=1in,height=1.25in,clip,keepaspectratio]{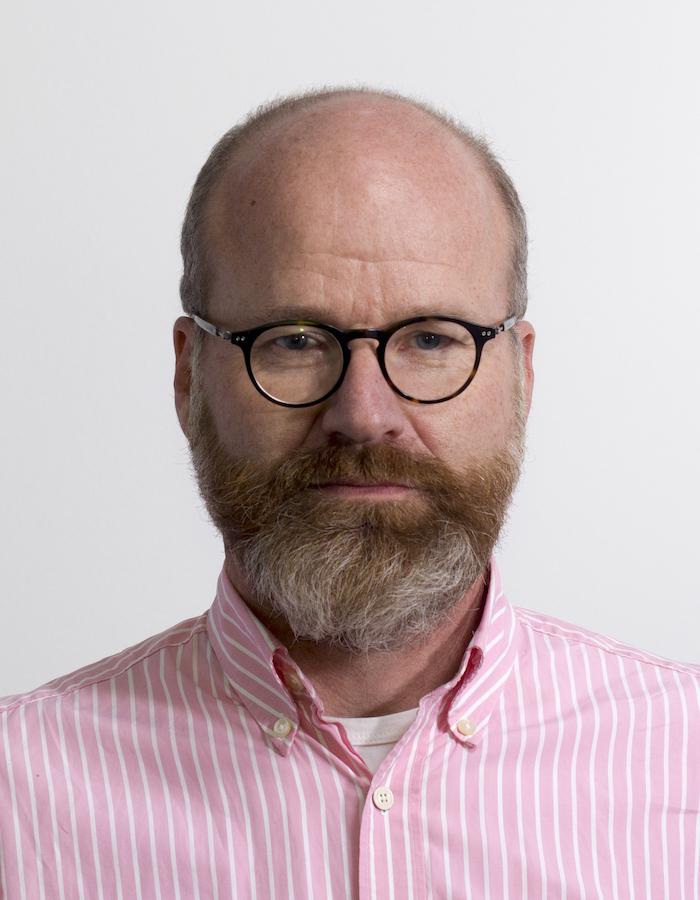}}]{Patrick van der Smagt}
received his Ph.D. degree in mathematics and computer science from the University of Amsterdam. He is director of AI Research at Volkswagen Group, head of the Volkswagen Group Machine Learning Research Lab in Munich, and holds a honorary professorship at ELTE University Budapest. He previously directed a lab as professor for machine learning and biomimetic robotics at the Technical University of Munich while leading the machine learning group at the research institute fortiss. Patrick van der Smagt has won numerous awards, including the 2013 Helmholtz-Association Erwin Schroedinger Award, the 2014 King-Sun Fu Memorial Award, the 2013 Harvard Medical School/MGH Martin Research Prize, the 2018 Webit Best Implementation of AI Award, and best-paper awards at machine learning and robotics conferences and journals.
\end{IEEEbiography}
\begin{IEEEbiography}[{\includegraphics[width=1in,height=1.25in,clip,keepaspectratio]{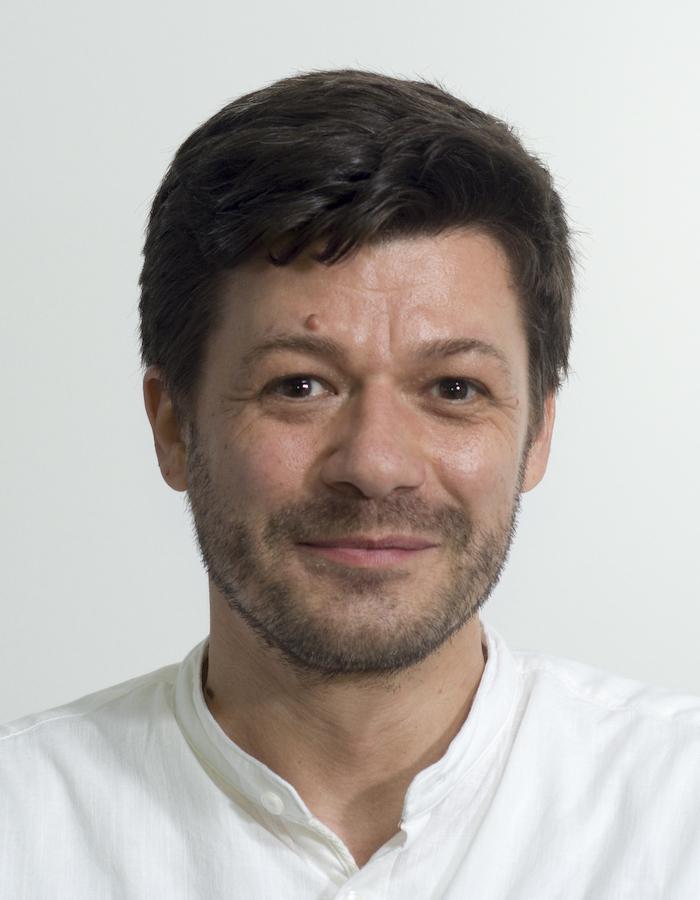}}]{Botond Cseke}
received his  Ph.D. degree  in computer science  from Radboud University Nijmegen. After post-doc positions at University of Edinburgh and Microsoft Research Cambridge, he joined the Volkswagen Group Machine Learning Research Lab in 2017 as a research scientist. He is interested in approximate probabilistic inference and related applications.
\end{IEEEbiography}

\end{document}